\documentclass{article}
\usepackage{amssymb}
\usepackage{todonotes}

\usepackage[preprint]{corl_2025} 

\title{FLARE: Robot Learning with \\ Implicit World Modeling}

\usepackage[normalem]{ulem}

\author{
  \textbf{Ruijie Zheng}\textsuperscript{1,2*}, \textbf{Jing Wang}\textsuperscript{1,3*}, \textbf{Scott Reed}\textsuperscript{1*} \\
  \textbf{Johan Bjorck}\textsuperscript{1\dag}, \textbf{Yu Fang}\textsuperscript{1\dag}, \textbf{Fengyuan Hu}\textsuperscript{1\dag}, \textbf{Joel Jang}\textsuperscript{1\dag}, \textbf{Kaushil Kundalia}\textsuperscript{1\dag} \\
  \textbf{Zongyu Lin}\textsuperscript{1\dag}, \textbf{Loic Magne}\textsuperscript{1\dag}, \textbf{Avnish Narayan}\textsuperscript{1\dag}, \textbf{You Liang Tan}\textsuperscript{1\dag}, \textbf{Guanzhi Wang}\textsuperscript{1\dag} \\
  \textbf{Qi Wang}\textsuperscript{1\dag}, \textbf{Jiannan Xiang}\textsuperscript{1\dag}, \textbf{Yinzhen Xu}\textsuperscript{1\dag}, \textbf{Seonghyeon Ye}\textsuperscript{1\dag} \\
  \textbf{Jan Kautz}\textsuperscript{1}, \textbf{Furong Huang}\textsuperscript{2}, \textbf{Yuke Zhu}\textsuperscript{1, 4\ddag}, \textbf{Linxi Fan}\textsuperscript{1\ddag}\\
  \textsuperscript{1}NVIDIA \quad
  \textsuperscript{2}University of Maryland, College Park \\
  \textsuperscript{3}Nanyang Technological University \quad
  \textsuperscript{4}University of Texas, Austin \\
  \textsuperscript{*}equal contribution \quad
  \textsuperscript{\dag}alphabetical order \quad
  \textsuperscript{\ddag}equal advising
}

\usepackage{pdfpages}
\usepackage{amsmath}
\usepackage{amsfonts}
\usepackage{mathtools}
\usepackage{amsthm}

\usepackage{microtype}
\usepackage{graphicx}
\usepackage{booktabs} 
\usepackage{algorithm}
\usepackage{algorithmic}
\usepackage{url}
\usepackage{tabularx}
\usepackage{graphicx}
\usepackage{bbm}
\usepackage{placeins}
\usepackage{adjustbox}
\usepackage{xcolor}
\usepackage{scalerel}
\usepackage{natbib}
\usepackage[outdir=./]{epstopdf}
\usepackage{graphicx,float,pgfplots,wrapfig,sidecap,lipsum}

\usepackage{colortbl}
\usepackage{diagbox} 
\usepackage{tablefootnote}
\usepackage[font=small,labelfont=bf]{caption}
\usepackage{subcaption}
\usepackage{subfloat}
\usepackage{tikz}
\usetikzlibrary{fit}
\usetikzlibrary{calc,shapes}
\usetikzlibrary{decorations.pathmorphing} 
\usetikzlibrary{fit}					
\usetikzlibrary{backgrounds}	
\usetikzlibrary{pgfplots.groupplots}

\usepackage[utf8]{inputenc}
\usepackage{pgfplots}
\usepgfplotslibrary{groupplots,dateplot}
\usetikzlibrary{patterns,shapes.arrows}
\pgfplotsset{compat=newest}

\usepackage{xspace}
\usepackage{soul}
\usepackage{booktabs}
\usepackage{bm}
	
\usepackage[capitalise]{cleveref}

\usepackage{listings}
\usepackage{cleveref}
 \usepackage{multirow}
\usepackage{xcolor}

\usepackage{makecell} 

\usepackage{amsmath,amsfonts,bm}

\def\eqref#1{equation~\ref{#1}}

\def\1{\bm{1}}

\DeclareMathAlphabet{\mathsfit}{\encodingdefault}{\sfdefault}{m}{sl}
\SetMathAlphabet{\mathsfit}{bold}{\encodingdefault}{\sfdefault}{bx}{n}

\theoremstyle{remark}

\AtBeginEnvironment{proof}{}{}{}

\definecolor{sourcecolor}{rgb}{0.5,1,0.5}
\definecolor{ourcolor}{rgb}{1,0,0}
\definecolor{singlecolor}{rgb}{0,0,1}
\definecolor{auxcolor}{rgb}{0.54,0.17,0.89}
\definecolor{linearcolor}{rgb}{0.172549019607843,0.627450980392157,0.172549019607843}
\definecolor{randomcolor}{rgb}{1,0.498039215686275,0.0549019607843137}
\definecolor{tunecolor}{rgb}{0.9568627450980393, 0.8156862745098039,0}
\definecolor{aligncolor}{rgb}{0,0.5,0}

\definecolor{LightGray}{gray}{0.9}

\definecolor{keywordcolor}{rgb}{0.0, 0.0, 0.8}  
\definecolor{commentcolor}{rgb}{0.0, 0.5, 0.0}  
\definecolor{stringcolor}{rgb}{0.58, 0.0, 0.82} 
\definecolor{backgroundcolor}{rgb}{1.,1.,1.} 

\newcommand{\modelname}{\textbf{\textsc{FLARE}}\xspace}

\begin{document}
\maketitle
\vspace{-3.0em}
\begin{center}
\url{https://research.nvidia.com/labs/gear/flare}
\end{center}

\begin{abstract}
We introduce \textit{\textbf{F}uture \textbf{LA}tent \textbf{RE}presentation Alignment (\modelname)}, a novel framework that integrates predictive latent world modeling into robot policy learning.
By aligning features from a diffusion transformer with latent embeddings of future observations, \modelname~enables a diffusion transformer policy to anticipate latent representations of future observations, allowing it to reason about long-term consequences while generating actions.
Remarkably lightweight, \modelname requires only minimal architectural modifications---adding a few tokens to standard vision-language-action (VLA) models---yet delivers substantial performance gains.
Across two challenging multitask simulation imitation learning benchmarks spanning single-arm and humanoid tabletop manipulation, \modelname achieves state-of-the-art performance, outperforming prior policy learning baselines by up to 26\%.
Moreover, \modelname unlocks the ability to co-train with human egocentric video demonstrations without action labels, significantly boosting policy generalization to a novel object with unseen geometry with as few as a single robot demonstration.
Our results establish \modelname as a general and scalable approach for combining implicit world modeling with high-frequency robotic control.

\end{abstract}

\keywords{World Model, VLA, Humanoid Robotics} 


\section{Introduction}
Human cognitive processes involve sophisticated predictive capabilities that operate largely implicitly. Consider a common action such as reaching for a coffee mug on a cluttered desk: without thinking about it, human brains could predict how the hand will move, what obstacles it might encounter, and how the mug will feel when grasped.
This capacity to construct internal representations of future states, a form of world modeling, is fundamental to efficient human motor control and decision-making. \looseness=-1


Several recent works~\citep{wu2024gr1, cheang2024gr2, li2025uva, zhu2025uwm, zhao2025cotvlavisualchainofthoughtreasoning, du2023unipi} have explored jointly learning world models and policies by generating future visual frames in parallel with actions.
While intuitive, this approach faces notable practical and conceptual challenges.
High-fidelity visual prediction typically requires large-scale generative models, introducing significant computational overhead and latency. 
Moreover, optimizing simultaneously for pixel-level reconstruction and action prediction places competing demands on model capacity:
visual generation emphasizes detailed spatial fidelity and texture synthesis, whereas action modeling benefits from compact, abstract, task-relevant representations, often leading to diluted learning efficiency.
In this work, we show that a surprisingly simple and flexible recipe, fully compatible with existing VLA architectures, can surpass prior VLA policy learning methods by a substantial margin.

\begin{figure}[t]
\centering
\includegraphics[trim={2cm 0cm 2cm 0}, width=0.95\linewidth]{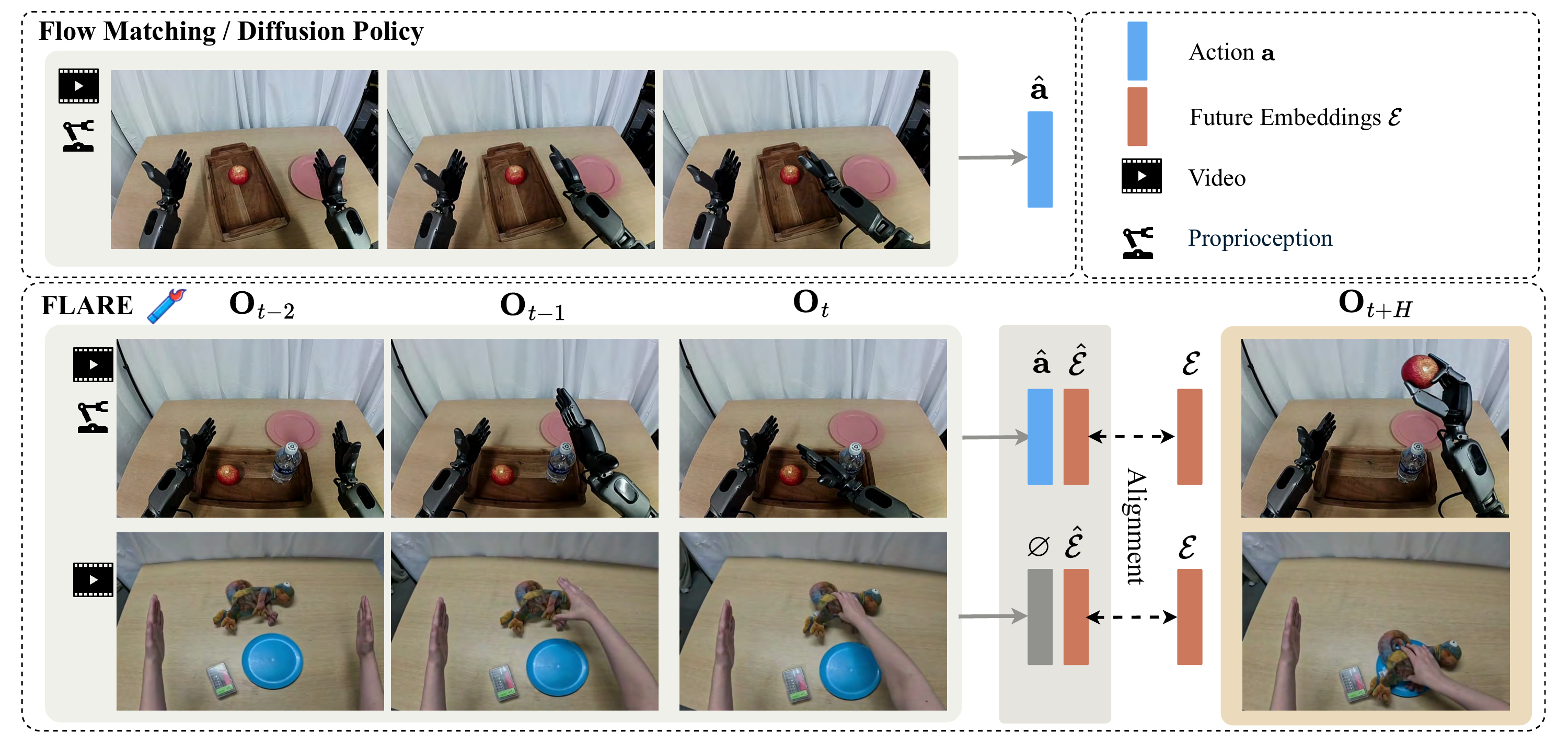}
\caption{Comparison of \modelname{} to a conventional flow-matching (or diffusion) policy. \modelname{} can train using both action flow-matching and future latent alignment objectives, leading to improved performance as well as enabling learning from video-only data such as human ego-view demonstrations.}
\label{fig:flare_concept}
\end{figure}

We introduce \textbf{F}uture L\textbf{A}tent \textbf{RE}presentation Alignment (\textbf{\modelname}), a lightweight yet highly effective extension to diffusion or flow-matching policies that introduces latent-space world modeling via a future alignment objective, eliminating the need for full-frame reconstruction. 
At its core, \modelname predicts a compact representation of the robot's future observation from the hidden states of the action denoising network.
\modelname operates in two key stages. 
First, we pretrain a compact, action-aware observation embedding model. 
While general-purpose embedding models could be used for the target future embeddings, we find that an action-aware embedding explicitly optimized for downstream control tasks offers superior performance and efficiency due to its compactness and task alignment.
Next, we co-train the diffusion transformer by introducing a minimal set of additional tokens, which are optimized to predict the future observation embeddings.
This approach requires minimal modifications to existing VLA architectures~\citep{black2024pi_0, nvidia2025gr00tn1}, making it broadly applicable and easy to deploy. \looseness=-1

Despite its simplicity, \modelname~achieves state-of-the-art performance across two multitask imitation-learning benchmarks spanning single-arm and humanoid tabletop manipulation.
Notably, when trained on diverse cross-embodiment robot data, our action-aware embedding model generalizes effectively to unseen embodiment and tasks.
With just 100 trajectories per task collected on a real GR1 humanoid posttrained from our pretrained action-aware observation embedding model, the \modelname policy achieves a 95\% success rate in real-world evaluations.
Finally, \modelname~enables learning from action-free data sources, such as human videos. 
By leveraging GoPro-collected human egocentric video demonstrations and only a single real robot demonstration per object,  \modelname successfully learns novel grasping strategies, highlighting its potential for scalable robot learning from less structured data sources.





\vspace{-.05in}
\section{Background}
\vspace{-.02in}
In this work, following $\pi_0$ and GR00T N1~\citep{black2024pi_0, nvidia2025gr00tn1}, we adopt \textbf{flow-matching}~\citep{flowmatching} as the learning objective for fitting actions from human demonstrations. 
Let $o_t$ denote the robot’s observation, which includes image inputs (potentially from multiple views) and a language instruction; let $q_t$ be the robot’s proprioceptive state; and let $A_t = (a_t, \dots, a_{t+H})$ be an action chunk drawn from expert demonstrations. We define $\phi_t = VL(o_t)$ as the vision-language embedding of the observation.

Given the VL embedding $\phi_t$, an action chunk $A_t$, a flow-matching timestep $\tau \in [0, 1]$, and sampled noise $\epsilon \sim \mathcal{N}(\mathbf{0}, \mathbf{I})$, we construct the noised action chunk as:
\begin{equation*}
A_t^{\tau} = \tau A_t + (1 - \tau) \epsilon.
\end{equation*}
Then the model prediction $V_\theta(\phi_t, A_t^{\tau}, q_t)$ is trained to approximate the denoising direction $\epsilon - A_t$, by minimizing the following flow-matching loss:
\begin{equation}
\label{eq:fm_loss}
\mathcal{L}_{\textit{fm}}(\theta) = \mathbb{E}_{\tau} \left[\| V_\theta(\phi_t, A_t^{\tau}, q_t) - (\epsilon - A_t)\|^2\right].
\end{equation}
We sample the timestep $\tau$ from the distribution $p(\tau) = \text{Beta}\left(\frac{s - \tau}{s}; 1.5, 1\right)$ with $s = 0.999$ as in~\citet{black2024pi_0}.
At inference time, we generate action chunks via $K$-step denoising. We first sample an initial chunk $A_t^0 \sim \mathcal{N}(\mathbf{0}, \mathbf{I})$, and then apply forward Euler integration to iteratively refine it:
\begin{equation*}
A_t^{\tau + 1/K} = A_t^{\tau} + \frac{1}{K} V_\theta(\phi_t, A_t^{\tau}, q_t).
\end{equation*}
Following GR00T N1~\citep{nvidia2025gr00tn1}, we set $K=4$ throughout all of our experiments, and we use the same Diffusion Transformer (DiT) architecture~\cite{peebles2023scalable} for $V_\theta$ with alternating cross-attention and self-attention layers to condition on the robot's vision language embedding $\phi_t$.
\vspace{-0.05in}
\section{Method}
\vspace{-0.05in}
\begin{figure}[t]
    \centering
    \includegraphics[width=0.95\linewidth]{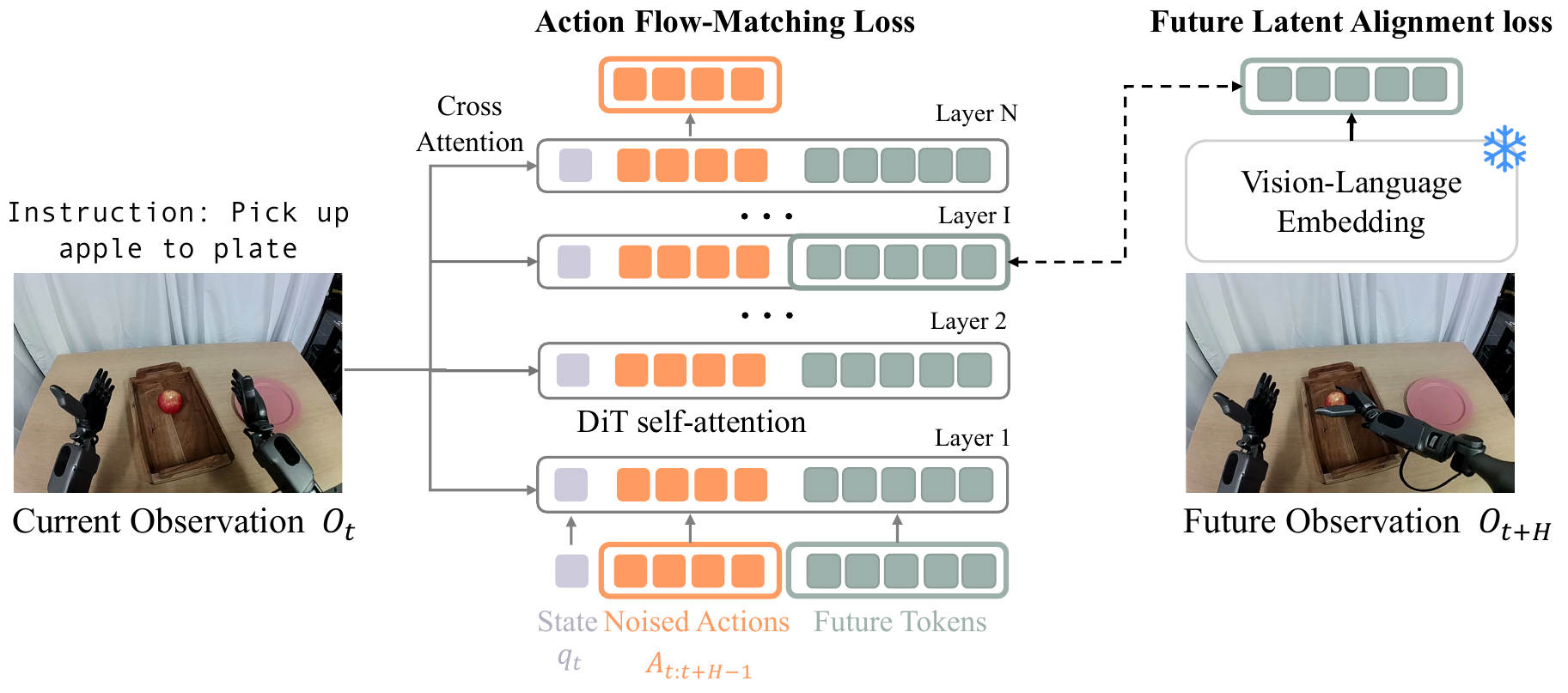}
    \caption{\modelname{} architecture. State and action token embeddings are concatenated into a sequence with learnable future token embeddings. The flow matching DiT blocks perform self-attention on this sequence, and cross-attention to the current vision and text observation embeddings. At a middle layer, the activations corresponding to the future token embeddings are used to compute a future latent alignment loss, which is the cosine similarity with vision-language embeddings from a future observation.}
    \label{fig:right_pdf}
    \vspace{-1.0em}
\end{figure}
\subsection{Latent World Modeling through Future Latent Representation Alignment}
\label{sec:joint_world}
To enable the latent representation within the DiT blocks to predict future latent states, we add $M$ learnable future token embeddings to the input sequence, such that the sequence contains three components: (1) the current proprioceptive state $q_t$ encoded via a state encoder, (2)  noised action chunk $A_t^\tau = \{\tau a_t + (1-\tau)\epsilon\}_{t}^{t+H}$ encoded by an action encoder, and (3) a set of $M$ learnable future tokens.
Next, we slice out the intermediate DiT representations corresponding to the $M$ future tokens at an internal layer $L$, project those features using an MLP, and finally align these with the frozen vision-language embeddings of the future observation $\phi_{t+H}$ (see Figure~\ref{fig:right_pdf}).

Our approach is similar to how Representation Alignment (REPA)~\citep{yu2025representation} is applied to improve text-to-image diffusion models, but with several important differences arising from the setting of latent world modeling. 
First, we align a DiT policy with \emph{future} embeddings, rather than embeddings of the current observation.
Second, our architecture adds learnable future tokens, so that the flow matching and alignment proceed along separate streams within the DiT, which interact via self-attention.

In this way, we encourage the DiT modules to internally reason about the future latent state while maintaining their action prediction capability through action flow-matching.
Letting $B$ indicate batch dimension and $D$ indicate embedding dimension, we can write the latent alignment objective as
\begin{align}
\label{eq:align_loss}
\mathcal{L}_{\textit{align}}(\theta) &= -\mathbb{E}_{\tau} \left[cos(f_{\theta}(\phi_t, A_{t}^{\tau}, q_t), g(\phi_{t+H})  \right] 
\end{align}
where $f_{\theta} \rightarrow \mathbb{R}^{B \times M \times D}$ outputs the DiT activations for the $M$ future tokens at layer $L$, and $g \rightarrow \mathbb{R}^{B \times M \times D}$ is the encoder of the future observation $\phi_{t+H}$.
The overall loss function is 
\begin{align}
\label{eq:flare_loss}
\mathcal{L} &= \mathcal{L}_{fm} + \lambda \mathcal{L}_{align} 
\end{align}
Empirically, we found $\lambda = 0.2$ worked the best in our experiments.
We refer the readers to Section~\ref{sec:ablation} for a detailed analysis of this choice.


\subsection{Action-aware Future Embedding Model}
\label{subsec:embedding_model}
While our future latent alignment framework is broadly compatible with various embedding models, we find that incorporating an \textit{action-aware} future embedding yields further improvements in both performance and efficiency.
To this end, we propose a compact vision-language embedding of the robot’s current observation, explicitly optimized for policy learning.
The design objective is twofold: achieving \textbf{compactness} while ensuring \textbf{action-awareness}.

Specifically, we leverage both the vision and text encoders from SigLIP-2~\citep{tschannen2025siglip} to encode the robot’s image observations and text instructions.
The encoded tokens are then fused using four layers of self-attention transformer blocks to capture cross-modal dependencies.
Subsequently, we apply a Q-former~\citep{blip2} module to compress the fused sequence into $M=32$ learnable query tokens, producing a compact, fixed-size representation that naturally generalizes to multi-camera inputs.
To ensure action-awareness, we train the vision language embedding end-to-end with the regular action flow-matching objective to predict the robot's actions by attaching 8 DiT blocks.
In this way, all task-relevant information is guaranteed to be captured within the latent token embeddings.
\begin{wrapfigure}{r}{0.51\textwidth}
    \centering
    \vspace{0pt}
    \includegraphics[width=\linewidth]{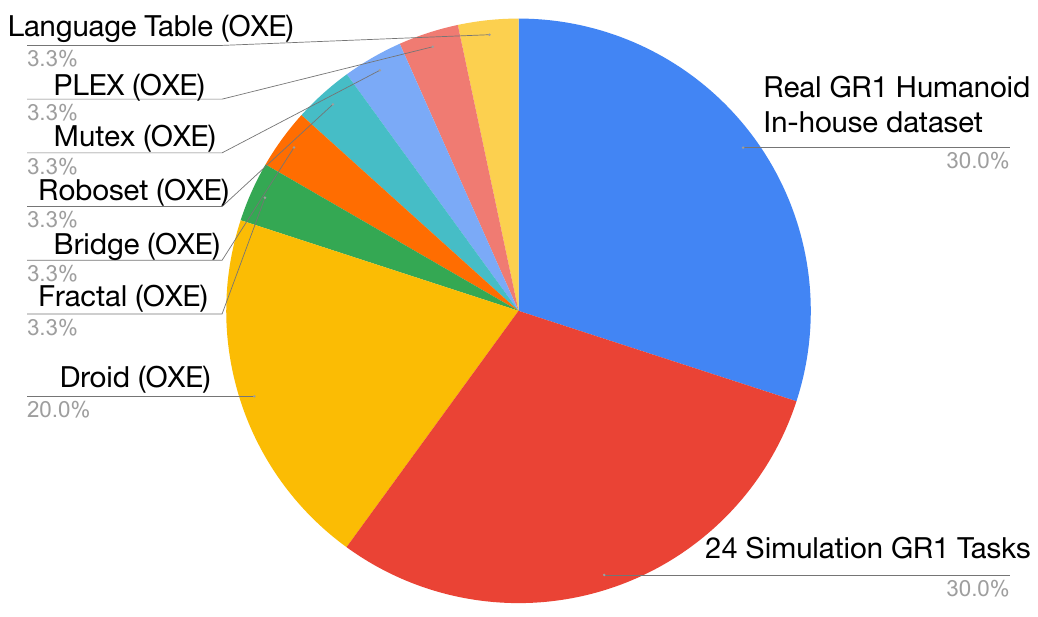}
    \caption{Data mixture of pretrained action-aware vision language embedding model}
    \label{fig:teacher_pretrain_mixture}
\end{wrapfigure}\\
To pretrain the embedding model, we leverage a diverse mixture of cross-embodiment robot datasets, comprising both simulated and real-world humanoid tabletop manipulation data from GR00T N1~\citep{nvidia2025gr00tn1} and seven additional datasets from Open X-Embodiment~\citep{open_x_embodiment_rt_x_2023}, totaling approximately 2,000 hours of robotic data.
Following pretraining, we posttrain the downstream policy jointly with the latent world model and the action prediction objective across downstream domains and tasks. 
Specifically, for posttraining, we initialize the downstream policy’s encoder with the pretrained embedding model, while also using the pretrained embedding model to define the prediction targets for future latent representations.
To mitigate distribution shifts between pretraining and downstream visual observations, rather than keeping the embedding model entirely frozen, we adopt an exponential moving average (EMA) update with respect to the policy's encoder. 
This strategy allows the embedding model to gradually adapt in tandem with the evolving vision and language encoders during policy fine-tuning.
Empirically, we find that an EMA update rate of 0.995 performs the best.
We refer the readers to Section~\ref{sec:ablation} for a detailed analysis of this choice.

\section{Experiments}
\begin{table*}[!thbp]
\centering
\small
\begin{tabularx}{\textwidth}{l*{5}{>{\centering\arraybackslash}X}}
\toprule
\textbf{Methods} & \textbf{FLARE} & \textbf{Policy Only} & \textbf{UWM} & \textbf{GR00T N1 (Scratch)} & \textbf{Diffusion Policy} \\
\midrule
Pick and Place & \textbf{53.2\%} & 43.8\% & 35.6\% & 44.1\% & 29.2\% \\
Open \& Close Doors / Drawers & \textbf{88.8\%} & 78.7\% & 82.0\% & 80.0\% & 78.7\% \\
Others & \textbf{80.0\%} & 75.2\% & 74.2\% & 69.6\% & 61.3\% \\
\midrule
\textbf{24 RoboCasa Tasks Average} & \textbf{70.1\%} & 61.9\% & 60.8\% & 60.6\% & 51.7\% \\
\midrule
Pick and Place Tasks & \textbf{58.2\%} & 46.6\% & 30.1\% & 51.8\% & 40.4\% \\
Articulated Tasks & \textbf{51.3\%} & 47.4\% & 38.4\% & 42.8\% & 50.1\% \\
\midrule
\textbf{24 GR1 Tasks Average} & \textbf{55.0\%} & 44.0\% & 29.5\% & 45.1\% & 40.9\% \\
\bottomrule
\end{tabularx}
\caption{Task Success Rate Breakdown for Multitask Policy on RoboCasa and GR1 Tabletop Manipulation}
\label{tab:combined_grouped_results}
\end{table*}
\begin{figure}[!htbp]
    \centering
    \includegraphics[width=1.0\linewidth]{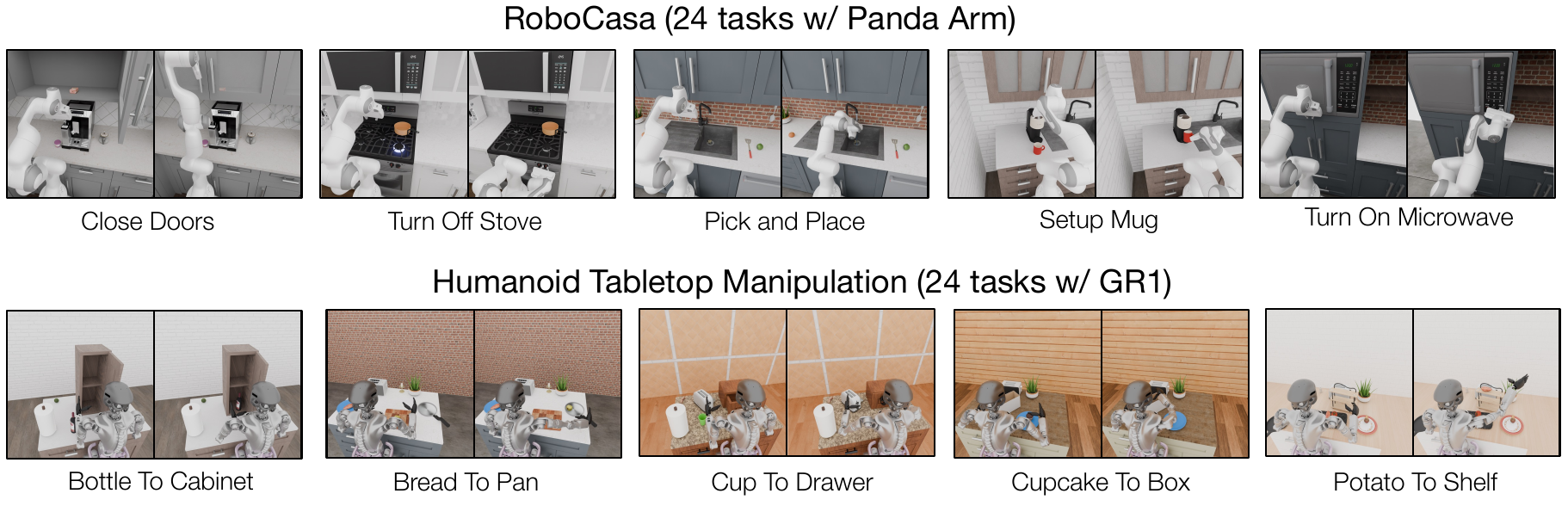}
    \caption{\textbf{Multitask Simulation Benchmarks}: We use 24 RoboCasa~\citep{robocasa2024} and 24 GR-1 tabletop manipulation tasks as a multitask simulation benchmark suite in this paper.}
    \vspace{-1.0em}
    \label{fig:task_fig}
\end{figure}
\subsection{Multitask Benchmark Performance}
\label{subsec:multitask}
In this section, we evaluate our latent world model on two multitask benchmarks that cover both single-arm manipulation and bimanual humanoid tabletop manipulation tasks. 
For the single-arm manipulation benchmark, we adopt RoboCasa~\citep{robocasa2024}, consisting of 24 atomic tasks in a simulated kitchen environment, including pick-and-place, door manipulation, faucet operation, and more. 
Robot's observations include three RGB images captured from cameras mounted on the left, right, and wrist of the robot.
Next, we incorporate 24 GR-1 tabletop simulation tasks from GR00T N1~\citep{nvidia2025gr00tn1}, which emphasize dexterous hand control with the GR-1 humanoid robot. 
This suite includes 18 object rearrangement tasks—picking up and placing objects between source and target containers—and 6 tasks involving interaction with articulated objects such as cabinets, drawers, and microwaves. 
Observation consists of a single RGB image from an egocentric camera positioned on the robot’s head. \looseness=-1

To ensure a fair comparison between our method and the baseline, for experiments in this section, we do not use the pretrained embedding model mentioned in Section~\ref{subsec:embedding_model}. 
Instead, we pretrain the embedding model exclusively on the same in-domain multitask dataset for 80,000 gradient steps, ensuring that any performance gains cannot be attributed to pretraining data with the embedding model.
In particular, we include the following baselines for the experimental results:
\begin{itemize}
    \item[1.]{Diffusion Policy~\citep{chi2024diffusionpolicy}}: Diffusion Policy models action distributions via a diffusion-based generative process, rather than using flow matching. It uses a U-Net architecture that progressively denoises random noise to generate the final action.
    \item[2.]{UWM~\citep{zhu2025uwm}}: We select UWM as the main baseline for methods that jointly learn video and action prediction objectives. UWM predicts image VAE latents and actions jointly with a diffusion objective.
    \item[3.]{GR00T N1 (Scratch)~\citep{nvidia2025gr00tn1}}: Since GR00T N1 is pretrained on a much broader data mixture, we ensure a fair comparison by using the same architecture but initializing the DiT layers from scratch, while only loading the pretrained Eagle VLM~\citep{eagle2} model weights.
    \item[4.]{\modelname~with Policy Only}: We use the exact same model architecture as \modelname, as mentioned in Section~\ref{subsec:embedding_model}, but train it solely with the policy learning objective. 
\end{itemize}
All methods are trained for 80,000 gradient steps on the multitask robot dataset, except for UWM.
We noticed that UWM performance is still improving at the end of 80k gradient steps, and thus we extend its training to 400k steps—five times the training budget allocated to the other methods.
Following GR00T N1~\cite{nvidia2025gr00tn1}, we evaluate each model checkpoint for 50 episodes per task every 1000 gradient steps, and report the maximum success rate over the final five checkpoints for each method.

As shown in Table~\ref{tab:combined_grouped_results}, we draw two key observations. 
First, \modelname{} consistently outperforms all baseline methods, including both the policy-only baselines and UWM. 
This highlights the strength of our compact, action-aware latent world modeling objective in enabling more effective policy learning.
Additionally, in our experiments, we also observe that~\modelname~with the policy-only objective, trained for 160k gradient steps, achieves only 44.1\% success rate, resulting in no performance difference compared with 80k gradient steps. 
Thus, the improved results cannot simply be attributed to more training steps with~\modelname. 
Second, even when trained with only the policy objective, FLARE still achieves performance on par with GR00T N1 initialized from scratch, despite GR00T N1 using a larger VLM backbone. This result underscores the quality of our Q-former-based vision-language embedding model in capturing action-relevant information.

\begin{figure}[!t]
    \centering
    \includegraphics[width=0.8\linewidth]{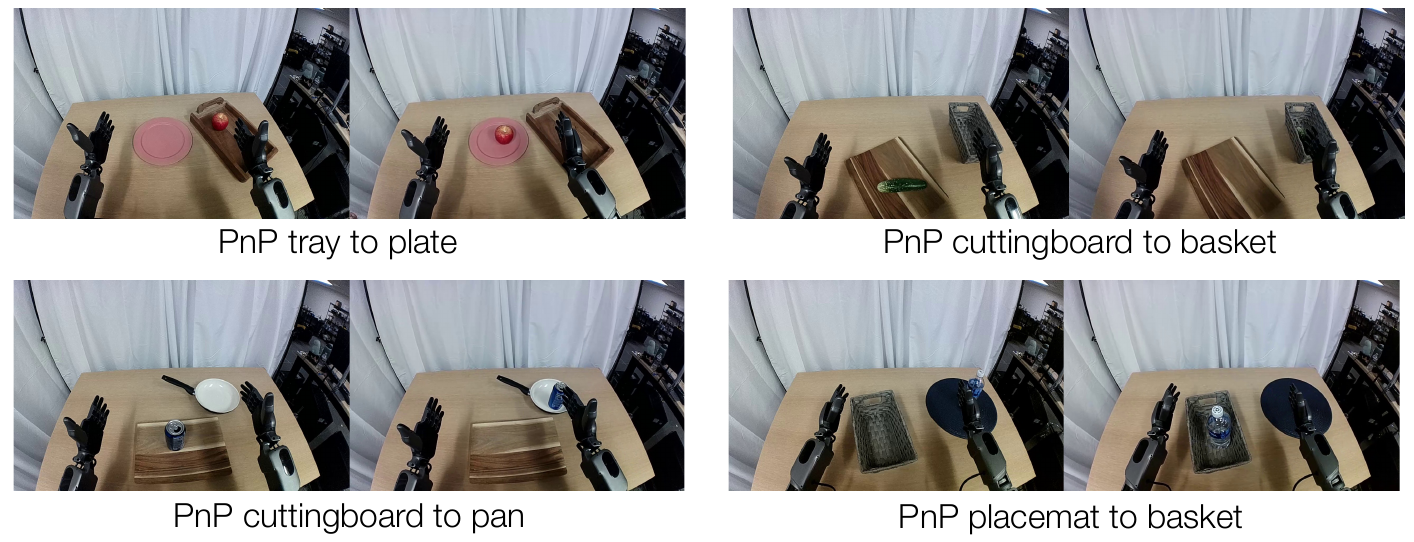}
    \caption{\textbf{Real GR1 Tasks Setup}: We evaluate four tabletop manipulation tasks on a real GR1 humanoid robot.}
    \label{fig:real_gr1_task}
\end{figure}
\begin{figure}[t]
    \centering
    \includegraphics[width=0.95\linewidth]{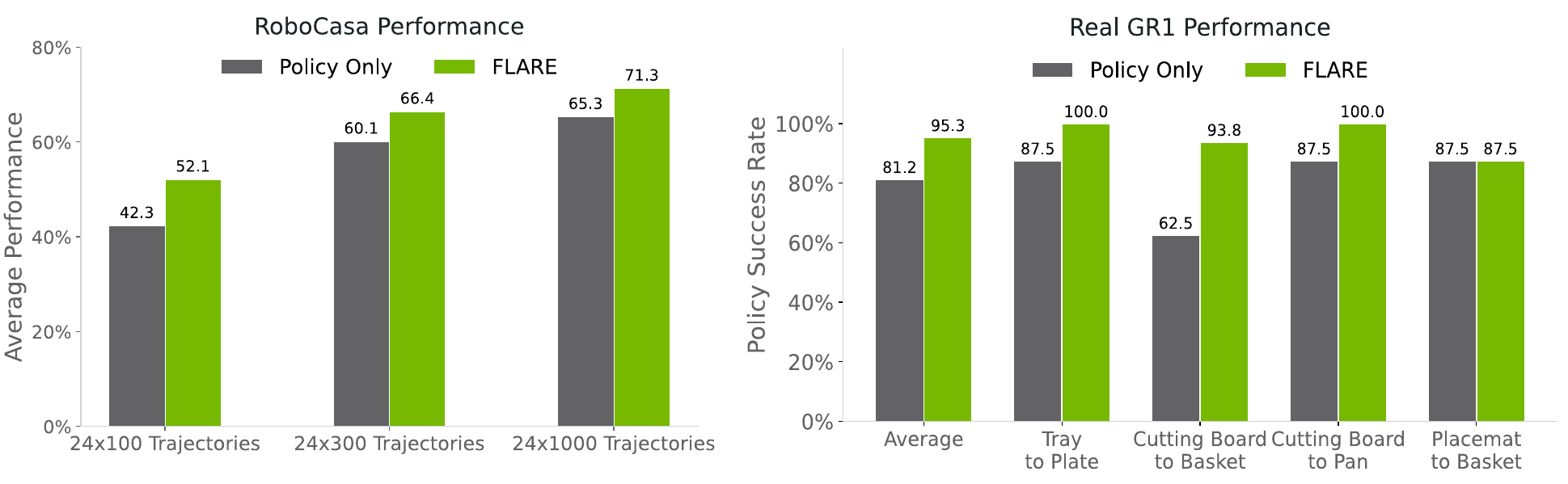}
    \caption{(\textbf{Left)}: Post-training results on 24 RoboCasa tasks.~\textbf{(Right)}: Post-training results on 4 Real GR1 humanoid tasks.}
    \label{fig:posttrain_results}
\end{figure}

\vspace{1.0em}
\subsection{Data-efficient Post-training with Cross-embodiment Pretrained Embedding Model}
\label{subsec:posttrain}
While the latent world model demonstrates substantial performance gains, as shown in the previous section, it requires training a separate embedding model for each domain.
In this section, we evaluate \modelname~with the pretrained embedding model mentioned in Section \ref{subsec:embedding_model} as the future prediction target, focusing on unseen embodiments and tasks with data-limited posttraining settings.
Specifically, we select 24 RoboCasa arm tasks and 4 real-world GR1 humanoid tabletop manipulation tasks as the evaluation benchmarks, and post-train the policy jointly with the latent world model and policy objectives, comparing it against a baseline that is post-trained using only the policy objective.
In particular, for the policy-only baseline, we initialize both the Q-former-based vision language embedding and the policy's DiT model weights from the cross-embodiment pretrained model.
For~\modelname, we only warm start the vision language embedding model.

For the evaluation protocol, we follow the same procedure described in Section~\ref{subsec:multitask} for the 24 RoboCasa tasks.
For the 4 real-world GR-1 tasks shown in Figure~\ref{fig:real_gr1_task}, we define 8 reference initial frames per task, each involving 4 distinct objects (apple, can, bottled water, cucumber) to manipulate, and report the success rate of the final policy checkpoint for each method.

As shown in Figure~\ref{fig:posttrain_results}, across both the 24 RoboCasa simulation tasks and the real-world GR-1 humanoid tasks, \modelname~consistently outperforms the policy-only baseline.
The improvement is especially pronounced under limited data conditions, achieving a 10\% gain on RoboCasa with 100 trajectories per task for posttraining.
Notably, although the pretrained embedding model has never seen RoboCasa tasks during pretraining, using it as the future embedding achieves comparable performance with 1000 trajectories to an embedding model trained exclusively on the 24 RoboCasa arm tasks (71.3\% vs.\ 70.2\% as reported in Section~\ref{subsec:multitask}). 

On the real GR-1 humanoid robot, we achieve a success rate of up to \textbf{95.1\%}, averaging 14\% higher than the baseline method.
Qualitatively, we observe that in scenarios where a can or water bottle is placed close to the robot’s hand, the baseline method trained with only the policy objective often knocks over the object.
In contrast, \modelname~policy learns to maneuver around or over the object and successfully grasp, highlighting the benefits of future latent reasoning enabled by \modelname.

\subsection{Leveraging Human Egocentric Trajectories without Action Labels}
While our previous experiments demonstrate that the proposed future latent alignment objective significantly enhances policy performance when trained on action-labeled data, we further show that it can be naturally extended to trajectories without action annotations, such as human egocentric demonstrations. 
This setting is particularly attractive, as collecting human demonstrations is substantially more cost-effective and efficient than teleoperating a robot to execute the same tasks.

\begin{figure}[t]
    \centering
    \begin{subfigure}[b]{0.572\linewidth}
        \centering
        \includegraphics[width=\linewidth]{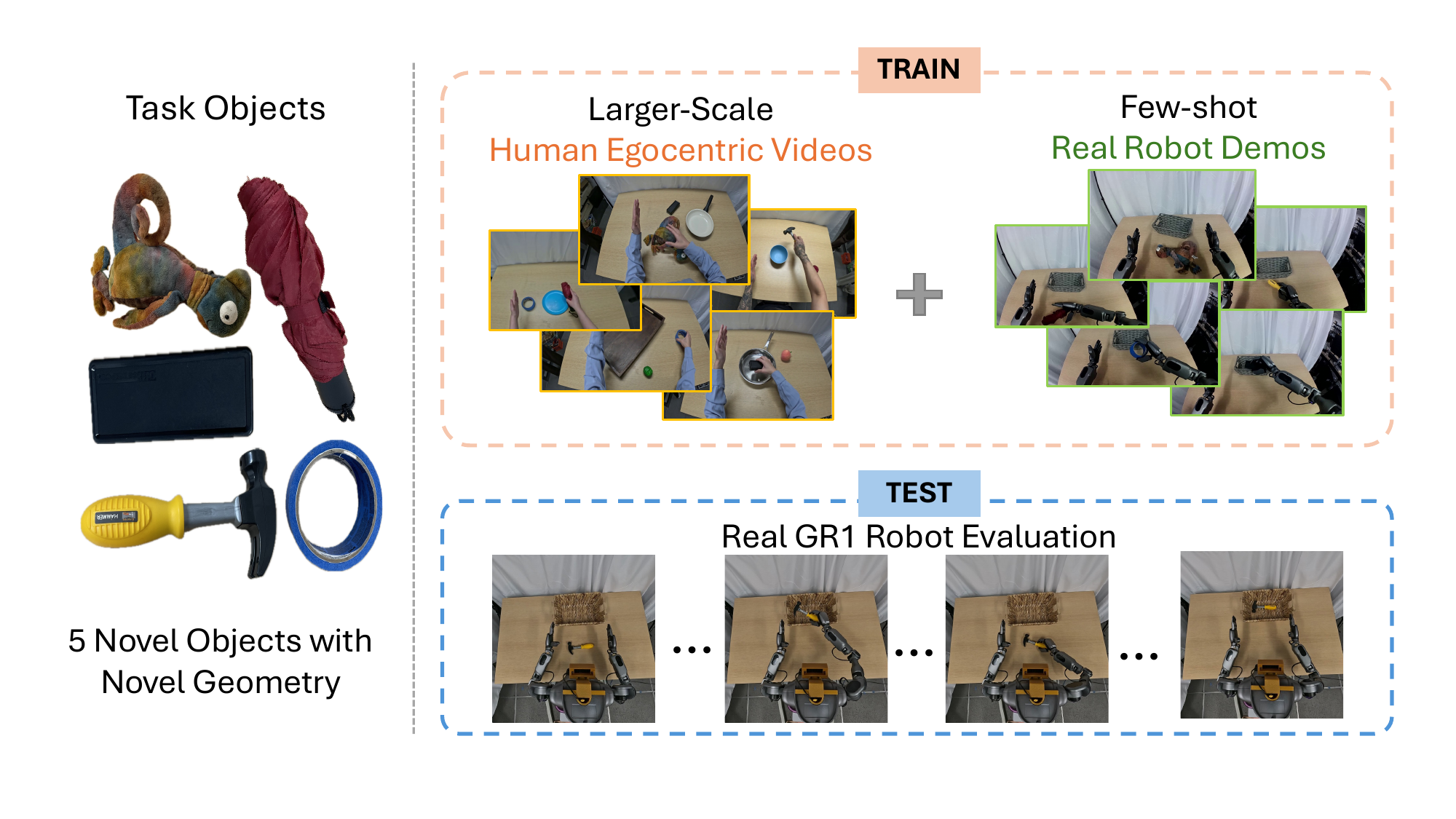}
        \label{fig:task_fig_a}
    \end{subfigure}
    \hfill
    \begin{subfigure}[b]{0.42\linewidth}
        \centering
        \includegraphics[width=\linewidth]{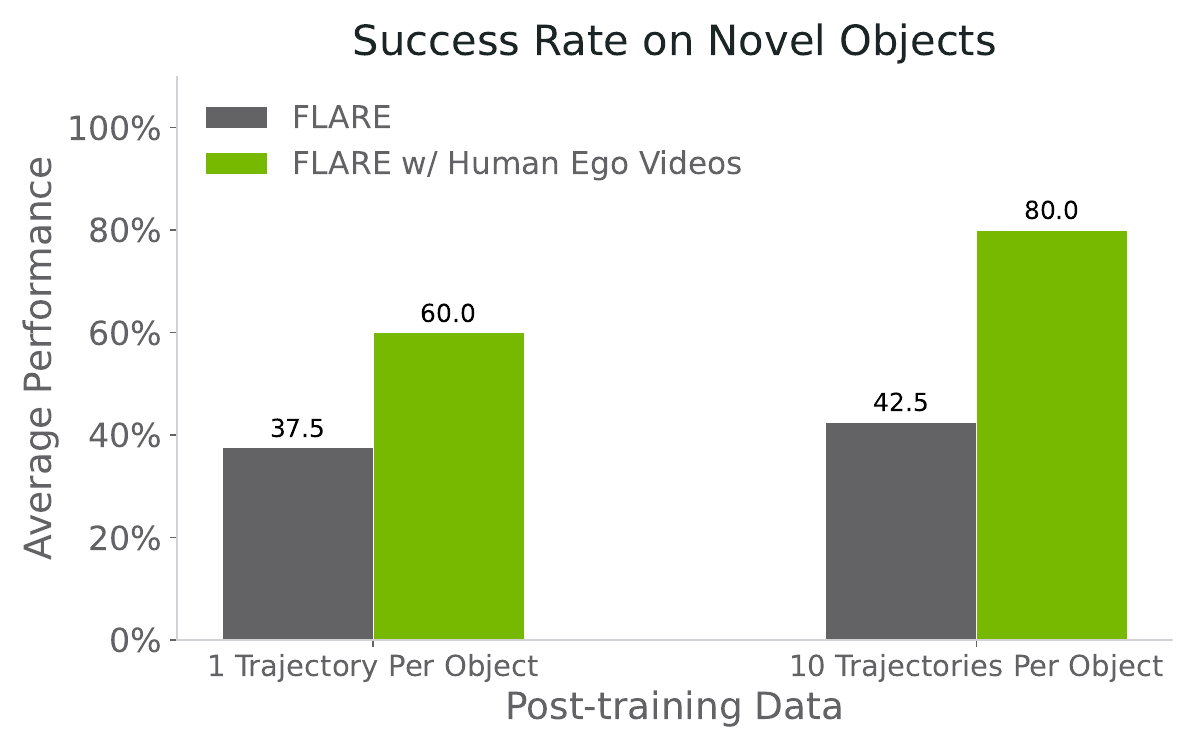}
        \label{fig:task_fig_b}
    \end{subfigure}
    \vspace{-3.0em}
    \caption{Generalizing to unseen objects with human egocentric videos and few-shot real robot demos}
    \label{fig:egocentric_results}
\end{figure}

To evaluate this, we select five novel objects with distinctive geometries that are absent from the training dataset, each requiring novel grasping strategies. For instance, the blue tape object is large and thus requires a top-down grasp by the robot hand. 
For each object, we collect 150 human egocentric demonstrations per object by mounting a GoPro on the demonstrator’s head while they perform similar tasks as the humanoid robot. 
On the robot side, we collect only 10 teleoperated demonstrations per object and train the policy using a mixture of these limited demonstrations, our GR-1 pretraining dataset, and the egocentric human videos.

For real-robot demonstrations with actions, we apply both the action flow-matching loss and the future alignment objective. In contrast, for the human egocentric videos without action labels, we rely solely on the future alignment loss to learn the latent dynamics. 
At evaluation time, we select five initial poses as reference images for each object and measure the robot's success rate. 
Partial credit (0.5) is given when the robot successfully grasps the object but fails to place it into the basket.

As shown in Figure~\ref{fig:egocentric_results}, with only \textbf{1} teleoperated trajectory per object, \modelname~already achieves up to a 60\% success rate on novel objects.
When provided with 10 trajectories per object, and jointly trained with human videos, \modelname~further improves to an 80\% success rate—roughly doubling the performance of a baseline trained solely on action-labeled data. 
These results highlight that \modelname~not only enhances learning from action-labeled demonstrations, but also effectively leverages unlabeled human demonstrations to improve generalization by capturing latent task dynamics.

\subsection{Ablation Study}
\label{sec:ablation}
\vspace{-0.5em}
\begin{wrapfigure}{r}{0.48\textwidth}
\vspace{-1em}
\begin{minipage}{0.48\textwidth}
\centering
\footnotesize
\begin{tabular}{l c}
\toprule
\textbf{Method} & \textbf{Success Rate (\%)} \\
\midrule
No FLARE loss & 43.9 \\
SigLIP2 & 49.6 \\
SigLIP2 (Average Pooled) & 50.9 \\
Action-aware Embedding & \textbf{55.0} \\
\bottomrule
\end{tabular}
\captionof{table}{Ablation of target embedding models.}
\label{tab:ablate_embedding}
\end{minipage}
\vspace{-10pt}
\end{wrapfigure}
\textbf{Using the Pretrained Siglip2 as Future Embedding model}:
While leveraging a policy-oriented future embedding model results in strong policy performance and enhanced training efficiency, we also explore an alternative setting that employs pretrained SigLIP2-Large vision tokens at timestep $t+16$ as prediction targets. 
Specifically, we experiment using both raw SigLIP2 vision tokens (256 tokens per image) and 2$\times$2 average-pooled tokens (64 tokens per image).
As illustrated in Table~\ref{tab:ablate_embedding}, our \modelname~framework maintains compatibility with diverse teacher encoder models beyond the policy-oriented embedding model. 
Although we get the optimal performance with the embedding model pretrained specifically on the target domain, using a more general-purpose vision encoder such as SigLIP2 still yields a significant 7\% improvement over baseline methods.
\begin{figure}[thbp!]
    \centering
    \includegraphics[width=0.8\linewidth]{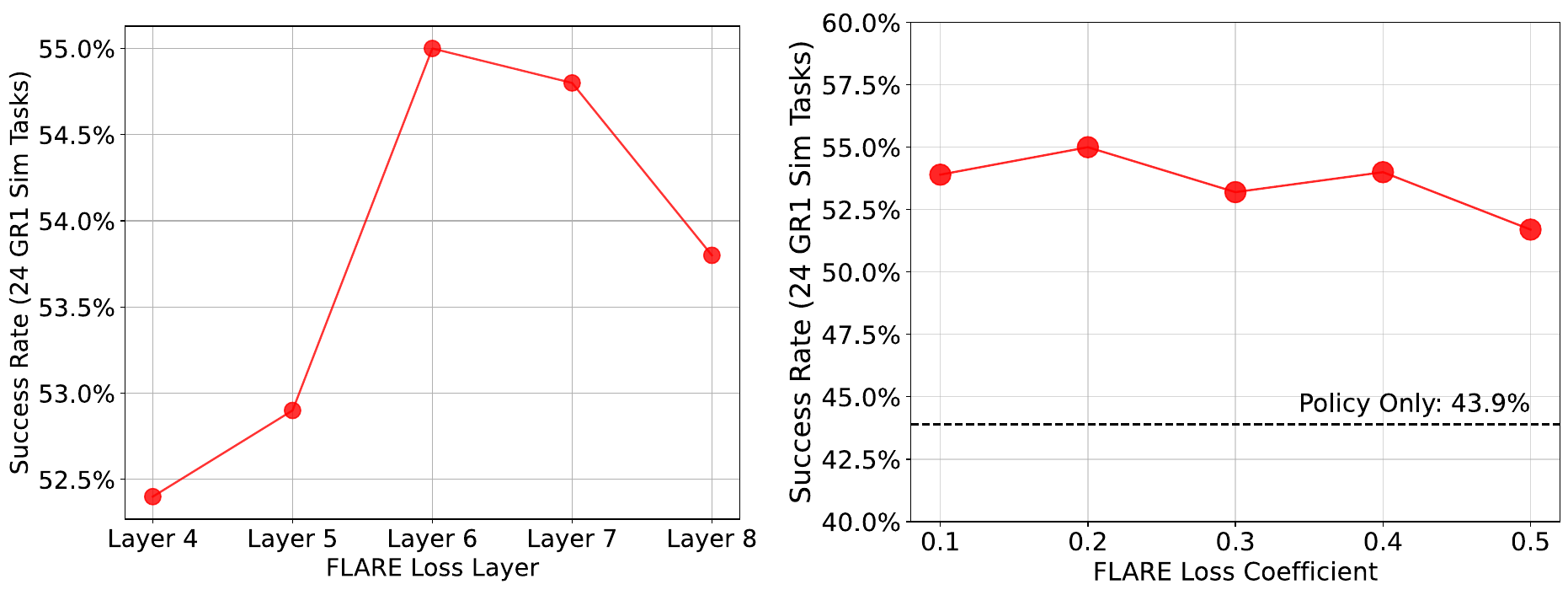}
    \caption{(\textbf{Left)}: Ablation of the DiT Layer used in FLARE loss (\textbf{Left)}: Ablation of FLARE loss coefficient.}
    \label{fig:ablation_results}
    \vspace{-1.0em}
\end{figure}

\textbf{Index of \modelname~Loss Layer and Coefficient of \modelname~Loss:}
A key design decision in \modelname~is selecting the DiT layer at which to apply the future latent alignment loss, and the coefficient $\lambda$ of \modelname~loss. 
In our main experiments, we apply this objective at layer 6 out of 8 total layers in the DiT architecture.
Applying it at deeper layers allows a larger portion of the model weights to benefit from the supervision of future latent prediction, but may also lead to conflicts between the action prediction and future alignment objectives.
To evaluate the effect of these two hyperparameters, we evaluate \modelname~on the GR1 simulation benchmark with different layer indexes and coefficients used for alignment.
As shown in Figure \ref{fig:ablation_results}, the model maintains strong performance across a range of hyperparameter setups. 
However, we do notice that applying the alignment objective too early—\textit{e.g.}, at layer 4—leads to a notable drop in performance, highlighting the importance of aligning the future prediction objective with the action denoising process.

\begin{figure}[!htbp]
    \centering
    \includegraphics[width=0.54\linewidth]{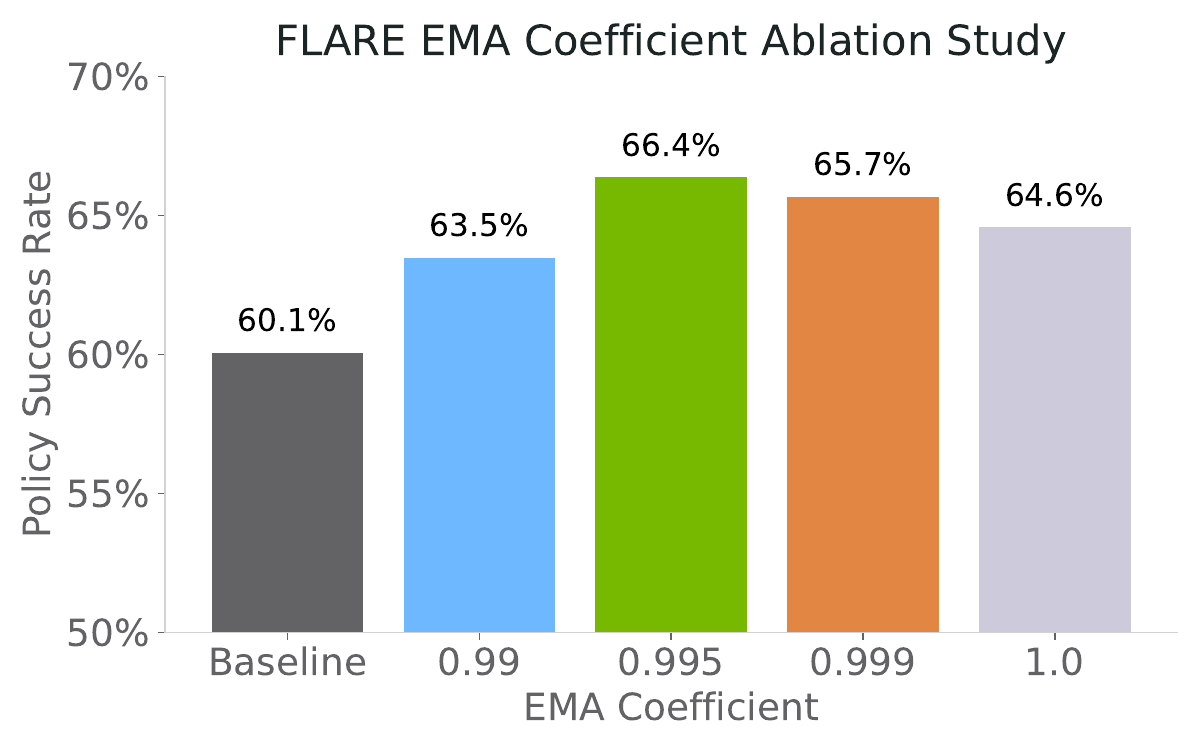}
    \caption{\textbf{Effect of EMA Coefficient $\rho$}: We report the policy success rate using $24\times 300$ training trajectories across 24 RoboCasa tasks. Baseline is trained without \modelname~future alignment loss, \textit{i.e.}, a policy-only objective.}
    \label{fig:ema}
\end{figure}

\textbf{Exponential Moving Average (EMA) of Pretrained Action-aware Embedding Model}:
As discussed in Section~\ref{subsec:embedding_model}, to address the distribution shift between pretraining and downstream tasks for our action-aware vision-language target embedding model, we incorporate an exponential moving average (EMA) update. 
Specifically, at each gradient step, the target embedding model parameters are updated as follows:
\begin{equation*}
\theta_{\text{target\_vl\_embedding}} \leftarrow \rho \theta_{\text{target\_embedding}} + (1-\rho) \theta_{\text{policy\_vl\_embedding}}
\end{equation*}

The EMA update enables the prediction target to adapt slowly in tandem with the evolving policy encoder, providing stability across training.
Here, We evaluate several choices of the EMA coefficient $\rho \in \{0.99, 0.995, 0.999, 1.0\}$, each using $24 \times 300$ trajectories to train the \modelname~policy. 
The final average success rates are reported in Figure~\ref{fig:ema}. 
We find that while all EMA variants outperform the baseline method without \modelname~future latent alignment objective, $\rho = 0.995$ yields the best performance and is used in all experiments. 
Notably, even with $\rho = 1.0$ (\textit{i.e.}, no EMA), \modelname~still surpasses the baseline, whereas $\rho = 0.99$ performs the worst, likely due to the instability caused by frequent target updates.

\section{Related Work}
\textbf{Generative World Models for Robotics}:
There has been a rich body of research on world models for robotics, ranging from model-based control to model-based reinforcement learning~\cite{jiang2020mbpo, hafner2020dreamerv2, Hansen2022tdmpc, cheng2025ramborlaugmentedmodelbasedoptimal, wang2023coplanner, zheng2023is}.
More recently, with advances in image and video generation, several works have explored the integration of generative modeling into policy learning~\citep{wu2024gr1, cheang2024gr2, du2023unipi, zhu2025uwm, li2025uva, du2024video}. 
One line of work~\citep{du2023unipi, huang2025ardupactiveregionvideo} uses image diffusion models with inverse dynamics models to close the perception-to-action loop.
The GR1 and GR2 families introduce end-to-end models that jointly predict discrete image tokens and actions using a unified next-token prediction objective. 
Other approaches~\citep{zhu2025uwm, li2025uva, zhou2024dino, pretrainspr2021, schwarzer2021spr, zheng2023taco, premiertaco2024} instead aim to jointly predict continuous image latents and actions.
For instance, UWM~\citep{zhu2025uwm} and UVA~\citep{li2025uva} jointly denoise VAE latents of future frames along with robot actions. 
DINO-WM~\cite{zhou2024dino} utilizes DINO features~\citep{zhou2024dino} to train a latent dynamics model for model-based planning.

Our work builds upon recent advances in representation learning, particularly Representation Alignment~\cite{yu2025representation}, which has shown remarkable success in accelerating the convergence of diffusion transformers for image generation and is key to state-of-the-art flow-matching models like Seedream-3.0~\citep{gao2025seedream}.
However, our approach differs in two crucial ways: we train a flow-matching \emph{policy} rather than an image model, and we align the DiT representation with features from \emph{future} observations rather than current ones.
In contrast to existing works, \modelname{} introduces an implicit latent world model objective that bypasses explicit reconstruction of future frames or latents.
This simple design enables reasoning over a compact, action-aware latent space and avoids the computational burden of high-fidelity generation, while maintaining compatibility with standard VLA architectures, without requiring major architectural redesign.
While DINO-WM focuses on zero-shot planning, \modelname{} is designed for policy and world model co-training, though planning could be a valuable future extension. \looseness=-1 

\textbf{Vision Language Action Models}.
A growing body of recent work~\citep{rt1-2022, rt22023arxiv, black2024pi_0, kim24openvla, zheng2025tracevla, wen2024tinyvla, cheang2024gr2vla, li2023vision, zhen20243dvla, huang2023embodied, ye2025latent, yang2025magma} has focused on developing general-purpose foundational vision-language-action (VLA) models by fine-tuning vision-language models for downstream robotics tasks.
Among these works, models such as~\citep{kim24openvla, yang2025magma, pertsch2025fastefficientactiontokenization, openvlaoft} autoregressively predict sequences of discrete action tokens using the next-token prediction objective.
In contrast, methods like~\citep{octo_2023, black2024pi_0, nvidia2025gr00tn1} leverage diffusion-based or flow-matching policy heads to bridge pretrained VLMs with continuous action generation.
In this work, inspired by the architecture of GR00T-N1~\citep{nvidia2025gr00tn1}, we adopt a flow-matching policy head built with diffusion transformer blocks, using interleaved self-attention and cross-attention layers to condition on the fused vision-language embeddings.

\textbf{Learning from Egocentric Videos}.
Several approaches have sought to enhance robot learning by leveraging human egocentric videos. These efforts extract diverse forms of information, such as human-object interactions~\citep{zeng2024learning}, object affordances~\citep{bahl2023affordances, kannan2023deft, srirama2024hrp, shaw2023videodex}, and visual trace trajectories~\citep{wen2023any, bharadhwaj2024track2act}. Other lines of work aim to translate human motions into robotic behaviors using hand pose estimators~\citep{wang2023mimicplay, zhu2024vision, shaw2023videodex, bharadhwaj2023zero, ye2023learning, qin2022dexmv} or motion capture systems~\citep{yang2024equibot}.
In this work, we show that future latent alignment provides a lightweight and effective alternative that does not require explicit pose estimators or point tracking tools, maximally reducing the engineering efforts.
A complementary direction focuses on learning latent actions from visual deltas between current and future frames to guide downstream policy learning~\citep{ye2025latent, bruce2024geniegenerativeinteractiveenvironments, chen2025motolatentmotiontoken, schmidt2024learning, ren2025videoworldexploringknowledgelearning, bu2025agibot}. 
Unlike latent actions as intermediate representations, whose correlation with ground-truth actions is unclear, our action-aware vision-language embedding directly aligns with future observations, resulting in a simple yet effective framework that naturally captures all the temporal dynamics information essential for effective policy learning. \looseness=-1


\section{Limitations}
In this work, we focus mainly on imitation learning with pick-and-place tasks on a real humanoid robot. 
Extending to more complex humanoid tasks that require more fine-grained dexterous manipulation, and incorporating reinforcement learning into the training paradigm, remains an important direction for future work.
Moreover, although our method enables generalization to novel objects, it still relies on a small number of expert demonstrations, which may limit scalability in settings where such data is hard to acquire.
Additionally, in this paper, we focus on egocentric human video datasets collected in controlled settings using head-mounted GoPro cameras. 
Extending to more diverse and larger-scale egocentric motion datasets captured in natural environments becomes a promising future direction of our work.

\section{Conclusion}
We present Future Latent Representation Alignment (\modelname), a simple yet effective framework for jointly learning robot policy and latent world dynamics. 
By aligning the future representations of the robot’s observations with the hidden states of the action denoising network, \modelname~enables the policy to implicitly reason about future states while predicting actions.
This approach leads to state-of-the-art performance on challenging robotic manipulation benchmarks. 
Furthermore, \modelname~unlocks co-training with human egocentric video demonstrations that lack action labels, significantly improving generalization to novel objects with minimal real-robot teleoperation data.
\subsection*{Acknowledgement}
We thank Jeremy Chimienti, Gianna Calderon, Isabel Zuluaga, Juan Zuluaga, Ivy Tam, Jazmin Sanchez, Jesse Yang, Leilee Naderi, Tri Cao for working with us on robot and GoPro teleoperation data collection and annotation.
This work is done during Ruijie Zheng and Jing Wang’s internship at NVIDIA. 
Zheng and Huang are supported by DARPA Transfer from Imprecise and Abstract Models to Autonomous Technologies (TIAMAT) 80321, National Science Foundation NSF-IIS-2147276 FAI, National Science Foundation NAIRR240045, DOD-AFOSR-Air Force Office of Scientific Research under award number FA9550-23-1-0048.





\clearpage
\appendix
\section{Details of Q-former based Vision Language Embedding Module}
\begin{figure}[!thbp]
    \centering
    \centering
    \includegraphics[width=0.8\linewidth]{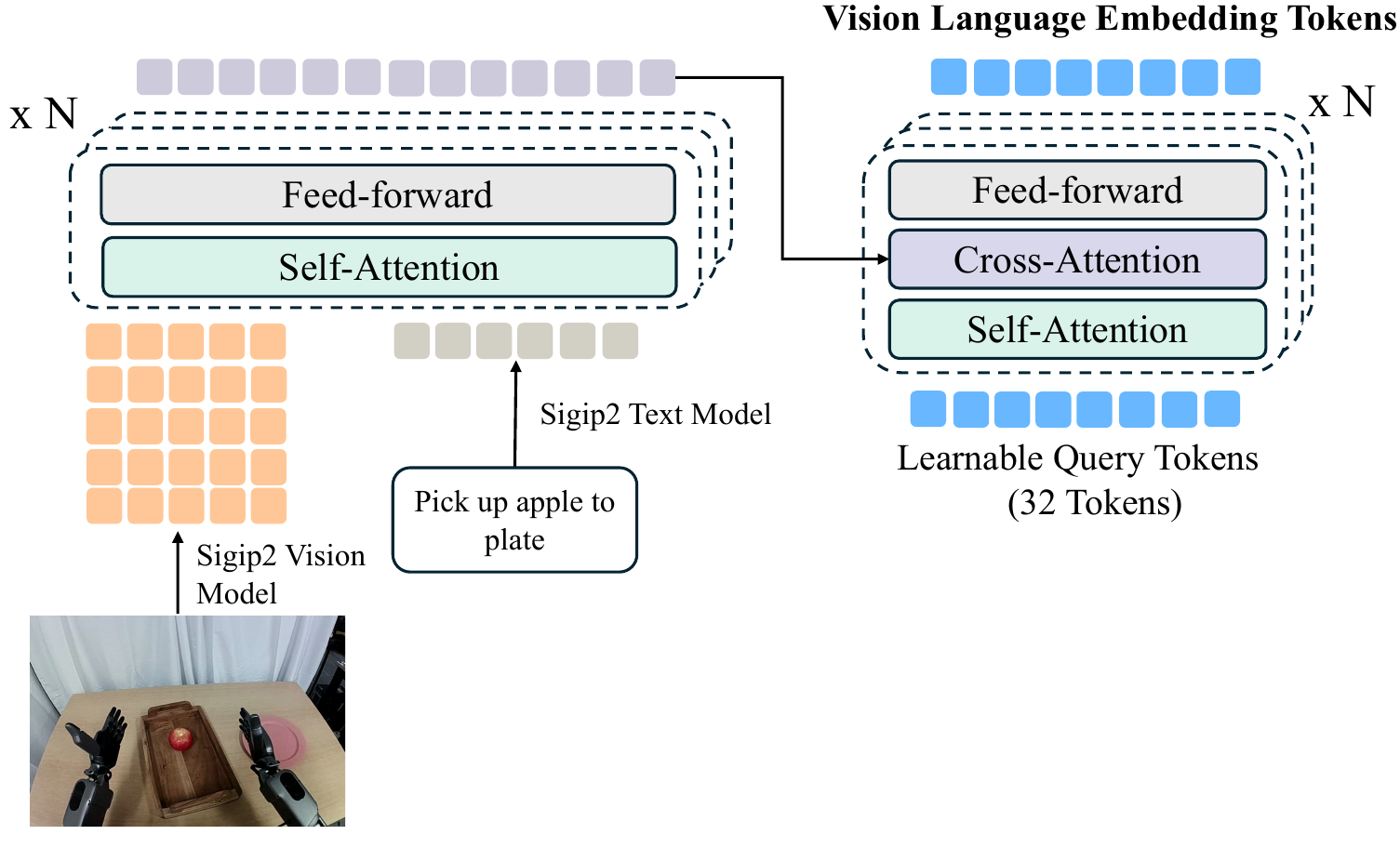}
    \caption{Our Q-former-based Vision Language Embedding Module}
    \label{fig:left_pdf}
\end{figure}
\label{sec:tokenizer}

We present the architectural details of our compact Q-former-based vision-language embedding module. 
Specifically, we adopt \texttt{siglip2-large-patch16-256} as the backbone for both vision and language encoders. 
The SigLIP2 vision encoder processes 256×256 resolution robot images into 256 patch tokens, while the language encoder encodes padded robot instructions into 32 language tokens. 
These 256 vision tokens and 32 language tokens are concatenated and passed through four layers of self-attention transformers to yield 288 fused vision-language tokens. 
To obtain a compact representation, we apply a Q-former architecture~\citep{blip2}, where 32 learnable query tokens—randomly initialized—interact with the 288 fused tokens through interleaved self-attention and cross-attention layers, producing 32 compressed vision-language tokens.

\section{Pretraining Data Mixture}
Details of pretraining data mixture are presented in Table~\ref{tab:dataset_stats}. 
\begin{table}[htbp]
\centering
\footnotesize
\caption{\centering Action-Aware Vision Language Embedding Pre-training Dataset Statistics}
\resizebox{\linewidth}{!}{%
\begin{tabular}{lrrrrl}
\toprule
\textbf{Dataset} & \textbf{Length (Frames)} & \textbf{Duration (hr)} & \textbf{FPS} & \textbf{Camera View} & \textbf{Category} \\ 
\midrule
GR-1 In-house Dataset & 6.4M & 88.4 & 20 & Egocentric & Real robot \\
DROID (OXE)~\citep{khazatsky2024droid} & 23.1M & 428.3 & 15 & Left, Right, Wrist & Real robot \\
RT-1 (OXE)~\citep{rt1-2022} & 3.7M & 338.4 & 3 & Egocentric & Real robot \\
Language Table (OXE)~\citep{lynch2022interactivelanguagetalkingrobots} & 7.0M & 195.7 & 10 & Front-facing & Real robot \\
Bridge-v2 (OXE)~\citep{walke2023bridgedata} & 2.0M & 111.1 & 5 & Shoulder, left, right, wrist & Real robot \\
MUTEX (OXE)~\citep{shah2023mutex} & 362K & 5.0 & 20 & Wrist & Real robot \\
Plex (OXE)~\citep{thomas2023plex} & 77K & 1.1 & 20 & Wrist & Real robot \\
RoboSet (OXE)~\citep{roboset} & 1.4M & 78.9 & 5 & Left, Right, Wrist & Real robot \\
GR-1 Simulation & 125.5M & 1,742.6 & 20 & Egocentric & Simulation \\
\midrule
Total & 169.5M & 2,989.5 & -- & -- & -- \\
\bottomrule
\end{tabular}
}
\label{tab:dataset_stats}
\end{table}

\section{Training Details}
For the pretraining of the action-aware vision language embedding module, we use 256 NVIDIA H100 GPUs with a batch size of 8192 for 150,000 gradient steps.
We use AdamW~\citep{adamw} optimizer with $\beta_1=0.95, \beta_2=0.999$, and $\epsilon=\text{1e-8}$. A weight decay of 1e-5 is applied, and the learning rate follows a cosine scheduling strategy with a warmup ratio of 0.05. 
Following~\citep{black2024pi_0, nvidia2025gr00tn1}, we sample the flowmatching denoising timestep from $p(\tau) = \text{Beta}(\frac{s-\tau}{s}; 1.5, 1)$, $s=0.999$.

For the multitask experiments of \modelname~conducted in Sections~\ref{subsec:multitask} and~\ref{subsec:posttrain}, we use 32 NVIDIA H100 GPUS with batch size 1024 for 80,000 gradient steps, while keeping the rest of the hyperparameter setups exactly the same.


\section{Pseudocode of \modelname}
Here we present a Python-style pseudocode of \modelname~loss calculation as well as the entire training loop. \looseness=-1

\begin{algorithm}[thb!]
   \caption{Python-style pseudocode for FLARE training}
   \label{alg:flare_pseudocode}
   
    \definecolor{codecomment}{rgb}{0.25,0.5,0.5}
    \definecolor{codekeyword}{rgb}{0.6,0.25,0.6}
    \lstset{
      basicstyle=\fontsize{7.2pt}{7.2pt}\ttfamily\bfseries,
      commentstyle=\fontsize{7.2pt}{7.2pt}\color{codecomment},
      keywordstyle=\fontsize{7.2pt}{7.2pt}\color{codekeyword},
      backgroundcolor=\color{white},  
    }
\begin{lstlisting}[language=python]
# target_vl_embedding: pretrained action-aware vision language embedding
# vl_embedding:  vision language embedding of the current policy
# dit: diffusion transformer of the current policy
# action_embedding: 2-layer MLP to embed noisy actions
# state_embedding:  2-layer MLP to embed prioprioceptive state
# action_decode:    2-layer MLP to decode robot's actions
# embedding_decode: 2-layer MLP to decode predicted embeddings
# N: Number of gradient steps
# M: Number of tokens in VL
# lambda: coefficient of FLARE loss (default is 0.2)

### Initialization
future_tokens = nn.Embedding(M, hiddem_dim)
vl_embedding.load_state_dict(vl_embedding.state_dict())
target_vl_embedding.requires_grad = False

for n in range(N): 
    obs, proprio, actions, future_obs = dataset.next()

    ### Prepare noisy action inputs
    noise = gaussian.sample()
    timestep = beta.sample() # sample flowmatching timestep
    noisy_action = timestep * actions + (1-timestep) * noise
    velocity = actions - noise

    ### Get state, action, and observation embedding tokens
    action_tokens = action_embed(noisy_action, timestep)
    state_token   = state_embed(state)
    vl_tokens     = vl_embedding(obs)
    
    ### Pass through DiT layers
    sa_tokens  = torch.concat([state_token, action_tokens, future_tokens], dim=1)
    policy_outputs = dit(sa_tokens, vl_tokens)

    ### Calculate action flowmatching loss
    action_outputs = action_decoder(policy_outputs[:, 1:1 + action_tokens.shape[1]])
    action_loss = MSE(action_outputs, velocity)

    ### Calculate FLARE loss
    with torch.no_grad():
        embedding_to_align = target_vl_embedding(future_obs)
    predict_embedding = decode_embedding(policy_outputs[:, -M:])    
    flare_loss = 1-COSINE_SIMILARITY(predict_embedding, embedding_to_align)

    ### Optimize the combined loss
    loss = action_loss + lambda * flare_loss
    optimizer.zero_grad()
    loss.backward()
    optimizer.step()

    
    
\end{lstlisting}
\label{alg: inference}
\end{algorithm}
\newpage
\section{Real GR1 Humanoid Rollouts}
\subsection{4 Pick-and-place Tasks}
Below, we present policy rollouts from the \modelname~trained policy on 4 real-world GR1 humanoid pick-and-place tasks, together with the task's language instructions.  
Qualitatively, we observe that when manipulating objects such as a bottled water or a Coke can, the \modelname~policy learns to maneuver the hand around the object, hovering over the water bottle, rather than striking and knocking it over.
\begin{figure}[!thbp]
    \centering
    \centering
    \includegraphics[width=1.0\linewidth]{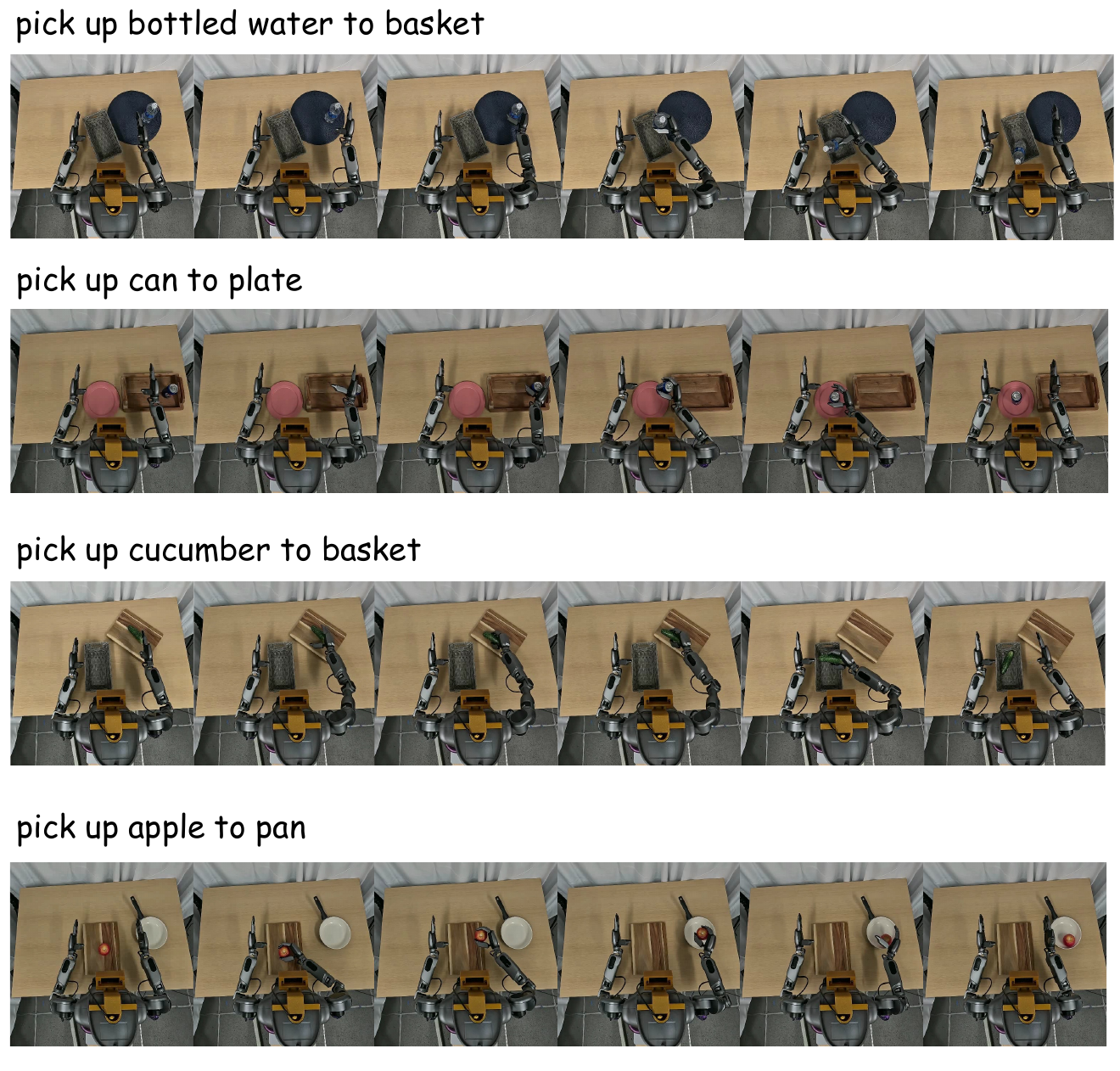}
    \caption{\modelname~policy rollout on real GR1 humanoid robot with 4 pick-and-place tasks}
    \label{fig:real_gr1_rollout}
\end{figure}

\newpage

\subsection{Manipulating Novel Objects}

Below, we present policy rollouts from the \modelname~trained policy manipulating 5 novel objects.  
\begin{figure}[!thbp]
    \centering
    \centering
    \includegraphics[width=1.0\linewidth]{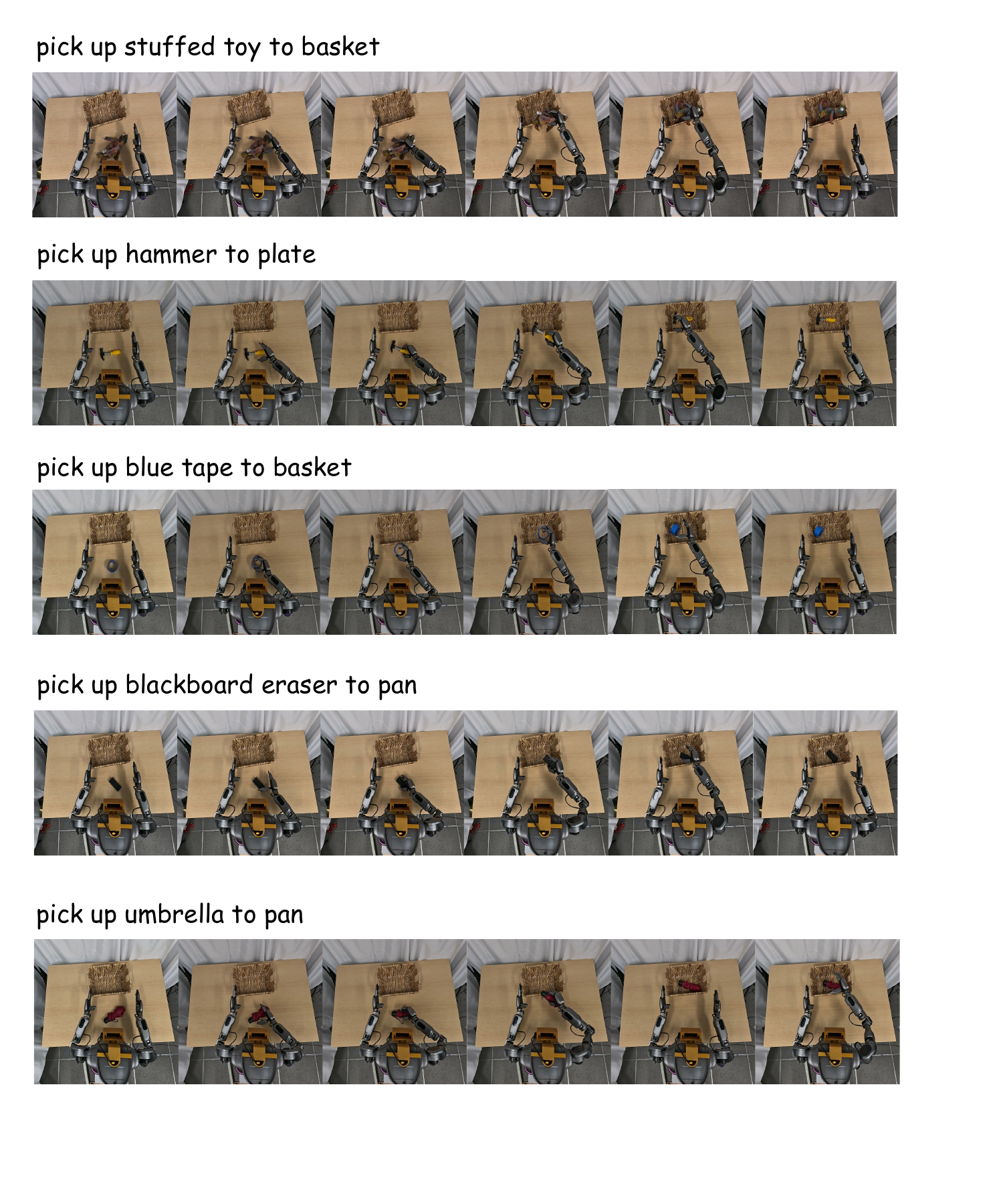}
    \caption{\modelname~policy rollout manipulating 5 novel objects}
    \label{fig:real_gr1_rollout_novel_object}
\end{figure}

\clearpage

\bibliography{citation}  

\begin{thebibliography}{70}
\providecommand{\natexlab}[1]{#1}
\providecommand{\url}[1]{\texttt{#1}}
\expandafter\ifx\csname urlstyle\endcsname\relax
  \providecommand{\doi}[1]{doi: #1}\else
  \providecommand{\doi}{doi: \begingroup \urlstyle{rm}\Url}\fi

\bibitem[Wu et~al.(2024)Wu, Jing, Cheang, Chen, Xu, Li, Liu, Li, and Kong]{wu2024gr1}
H.~Wu, Y.~Jing, C.~Cheang, G.~Chen, J.~Xu, X.~Li, M.~Liu, H.~Li, and T.~Kong.
\newblock Unleashing large-scale video generative pre-training for visual robot manipulation.
\newblock In \emph{The Twelfth International Conference on Learning Representations}, 2024.
\newblock URL \url{https://openreview.net/forum?id=NxoFmGgWC9}.

\bibitem[Cheang et~al.(2024)Cheang, Chen, Jing, Kong, Li, Li, Liu, Wu, Xu, Yang, Zhang, and Zhu]{cheang2024gr2}
C.-L. Cheang, G.~Chen, Y.~Jing, T.~Kong, H.~Li, Y.~Li, Y.~Liu, H.~Wu, J.~Xu, Y.~Yang, H.~Zhang, and M.~Zhu.
\newblock Gr-2: A generative video-language-action model with web-scale knowledge for robot manipulation.
\newblock \emph{arXiv preprint arXiv:2410.06158}, 2024.

\bibitem[Li et~al.(2025)Li, Gao, Sadigh, and Song]{li2025uva}
S.~Li, Y.~Gao, D.~Sadigh, and S.~Song.
\newblock Unified video action model.
\newblock \emph{arXiv preprint arXiv:2503.00200}, 2025.

\bibitem[Zhu et~al.(2025)Zhu, Yu, Feng, Burchfiel, Shah, and Gupta]{zhu2025uwm}
C.~Zhu, R.~Yu, S.~Feng, B.~Burchfiel, P.~Shah, and A.~Gupta.
\newblock Unified world models: Coupling video and action diffusion for pretraining on large robotic datasets.
\newblock 2025.

\bibitem[Zhao et~al.(2025)Zhao, Lu, Kim, Fu, Zhang, Wu, Li, Ma, Han, Finn, Handa, Liu, Xiang, Wetzstein, and Lin]{zhao2025cotvlavisualchainofthoughtreasoning}
Q.~Zhao, Y.~Lu, M.~J. Kim, Z.~Fu, Z.~Zhang, Y.~Wu, Z.~Li, Q.~Ma, S.~Han, C.~Finn, A.~Handa, M.-Y. Liu, D.~Xiang, G.~Wetzstein, and T.-Y. Lin.
\newblock Cot-vla: Visual chain-of-thought reasoning for vision-language-action models, 2025.
\newblock URL \url{https://arxiv.org/abs/2503.22020}.

\bibitem[Du et~al.(2023)Du, Yang, Dai, Dai, Nachum, Tenenbaum, Schuurmans, and Abbeel]{du2023unipi}
Y.~Du, S.~Yang, B.~Dai, H.~Dai, O.~Nachum, J.~B. Tenenbaum, D.~Schuurmans, and P.~Abbeel.
\newblock Learning universal policies via text-guided video generation.
\newblock In \emph{Thirty-seventh Conference on Neural Information Processing Systems}, 2023.
\newblock URL \url{https://openreview.net/forum?id=bo8q5MRcwy}.

\bibitem[Black et~al.(2024)Black, Brown, Driess, Esmail, Equi, Finn, Fusai, Groom, Hausman, Ichter, et~al.]{black2024pi_0}
K.~Black, N.~Brown, D.~Driess, A.~Esmail, M.~Equi, C.~Finn, N.~Fusai, L.~Groom, K.~Hausman, B.~Ichter, et~al.
\newblock $\pi_0$: A vision-language-action flow model for general robot control.
\newblock \emph{arXiv preprint arXiv:2410.24164}, 2024.

\bibitem[NVIDIA et~al.(2025)NVIDIA, :, Bjorck, Castañeda, Cherniadev, Da, Ding, Fan, Fang, Fox, Hu, Huang, Jang, Jiang, Kautz, Kundalia, Lao, Li, Lin, Lin, Liu, Llontop, Magne, Mandlekar, Narayan, Nasiriany, Reed, Tan, Wang, Wang, Wang, Wang, Xiang, Xie, Xu, Xu, Ye, Yu, Zhang, Zhang, Zhao, Zheng, and Zhu]{nvidia2025gr00tn1}
NVIDIA, :, J.~Bjorck, F.~Castañeda, N.~Cherniadev, X.~Da, R.~Ding, L.~J. Fan, Y.~Fang, D.~Fox, F.~Hu, S.~Huang, J.~Jang, Z.~Jiang, J.~Kautz, K.~Kundalia, L.~Lao, Z.~Li, Z.~Lin, K.~Lin, G.~Liu, E.~Llontop, L.~Magne, A.~Mandlekar, A.~Narayan, S.~Nasiriany, S.~Reed, Y.~L. Tan, G.~Wang, Z.~Wang, J.~Wang, Q.~Wang, J.~Xiang, Y.~Xie, Y.~Xu, Z.~Xu, S.~Ye, Z.~Yu, A.~Zhang, H.~Zhang, Y.~Zhao, R.~Zheng, and Y.~Zhu.
\newblock Gr00t n1: An open foundation model for generalist humanoid robots, 2025.
\newblock URL \url{https://arxiv.org/abs/2503.14734}.

\bibitem[Lipman et~al.()Lipman, Chen, Ben-Hamu, Nickel, and Le]{flowmatching}
Y.~Lipman, R.~T. Chen, H.~Ben-Hamu, M.~Nickel, and M.~Le.
\newblock Flow matching for generative modeling.
\newblock In \emph{The Eleventh International Conference on Learning Representations}.

\bibitem[Peebles and Xie(2023)]{peebles2023scalable}
W.~Peebles and S.~Xie.
\newblock Scalable diffusion models with transformers.
\newblock In \emph{Proceedings of the IEEE/CVF international conference on computer vision}, pages 4195--4205, 2023.

\bibitem[Yu et~al.(2025)Yu, Kwak, Jang, Jeong, Huang, Shin, and Xie]{yu2025representation}
S.~Yu, S.~Kwak, H.~Jang, J.~Jeong, J.~Huang, J.~Shin, and S.~Xie.
\newblock Representation alignment for generation: Training diffusion transformers is easier than you think.
\newblock In \emph{The Thirteenth International Conference on Learning Representations}, 2025.
\newblock URL \url{https://openreview.net/forum?id=DJSZGGZYVi}.

\bibitem[Tschannen et~al.(2025)Tschannen, Gritsenko, Wang, Naeem, Alabdulmohsin, Parthasarathy, Evans, Beyer, Xia, Mustafa, et~al.]{tschannen2025siglip}
M.~Tschannen, A.~Gritsenko, X.~Wang, M.~F. Naeem, I.~Alabdulmohsin, N.~Parthasarathy, T.~Evans, L.~Beyer, Y.~Xia, B.~Mustafa, et~al.
\newblock {SigLIP 2}: Multilingual vision-language encoders with improved semantic understanding, localization, and dense features.
\newblock \emph{arXiv preprint arXiv:2502.14786}, 2025.

\bibitem[Li et~al.(2023)Li, Li, Savarese, and Hoi]{blip2}
J.~Li, D.~Li, S.~Savarese, and S.~Hoi.
\newblock {BLIP}-2: Bootstrapping language-image pre-training with frozen image encoders and large language models.
\newblock In A.~Krause, E.~Brunskill, K.~Cho, B.~Engelhardt, S.~Sabato, and J.~Scarlett, editors, \emph{Proceedings of the 40th International Conference on Machine Learning}, volume 202 of \emph{Proceedings of Machine Learning Research}, pages 19730--19742. PMLR, 23--29 Jul 2023.
\newblock URL \url{https://proceedings.mlr.press/v202/li23q.html}.

\bibitem[{Open X-Embodiment Collaboration} et~al.(2024)]{open_x_embodiment_rt_x_2023}
{Open X-Embodiment Collaboration} et~al.
\newblock Open {X-E}mbodiment: Robotic learning datasets and {RT-X} models.
\newblock International Conference on Robotics and Automation, 2024.

\bibitem[Nasiriany et~al.(2024)Nasiriany, Maddukuri, Zhang, Parikh, Lo, Joshi, Mandlekar, and Zhu]{robocasa2024}
S.~Nasiriany, A.~Maddukuri, L.~Zhang, A.~Parikh, A.~Lo, A.~Joshi, A.~Mandlekar, and Y.~Zhu.
\newblock Robocasa: Large-scale simulation of everyday tasks for generalist robots.
\newblock In \emph{Robotics: Science and Systems (RSS)}, 2024.

\bibitem[Chi et~al.(2024)Chi, Xu, Feng, Cousineau, Du, Burchfiel, Tedrake, and Song]{chi2024diffusionpolicy}
C.~Chi, Z.~Xu, S.~Feng, E.~Cousineau, Y.~Du, B.~Burchfiel, R.~Tedrake, and S.~Song.
\newblock Diffusion policy: Visuomotor policy learning via action diffusion.
\newblock \emph{The International Journal of Robotics Research}, 2024.

\bibitem[Li et~al.(2025)Li, Chen, Liu, Wang, VS, Ji, Lan, Zhang, Zhao, Radhakrishnan, et~al.]{eagle2}
Z.~Li, G.~Chen, S.~Liu, S.~Wang, V.~VS, Y.~Ji, S.~Lan, H.~Zhang, Y.~Zhao, S.~Radhakrishnan, et~al.
\newblock Eagle 2: Building post-training data strategies from scratch for frontier vision-language models.
\newblock \emph{arXiv preprint arXiv:2501.14818}, 2025.

\bibitem[Jiang et~al.(2020)Jiang, Chen, Han, Li, Dong, and Zhang]{jiang2020mbpo}
X.~Jiang, Q.~Chen, S.~Han, M.~Li, J.~Dong, and R.~Zhang.
\newblock When to trust your model: Model-based policy optimization, 2020.
\newblock URL \url{https://openreview.net/forum?id=SkgPIpcGar}.
\newblock Submitted to NeurIPS 2019 Reproducibility Challenge.

\bibitem[Hafner et~al.(2020)Hafner, Lillicrap, Norouzi, and Ba]{hafner2020dreamerv2}
D.~Hafner, T.~Lillicrap, M.~Norouzi, and J.~Ba.
\newblock Mastering atari with discrete world models.
\newblock \emph{arXiv preprint arXiv:2010.02193}, 2020.

\bibitem[Hansen et~al.(2022)Hansen, Wang, and Su]{Hansen2022tdmpc}
N.~Hansen, X.~Wang, and H.~Su.
\newblock Temporal difference learning for model predictive control.
\newblock 2022.

\bibitem[Cheng et~al.(2025)Cheng, Kang, Fadini, Shi, and Coros]{cheng2025ramborlaugmentedmodelbasedoptimal}
J.~Cheng, D.~Kang, G.~Fadini, G.~Shi, and S.~Coros.
\newblock Rambo: Rl-augmented model-based optimal control for whole-body loco-manipulation, 2025.
\newblock URL \url{https://arxiv.org/abs/2504.06662}.

\bibitem[Wang et~al.(2023)Wang, Zheng, Sun, Jia, Wongkamjan, Xu, and Huang]{wang2023coplanner}
X.~Wang, R.~Zheng, Y.~Sun, R.~Jia, W.~Wongkamjan, H.~Xu, and F.~Huang.
\newblock {COP}lanner: Plan to roll out conservatively but to explore optimistically for model-based {RL}.
\newblock In \emph{NeurIPS 2023 Workshop on Generalization in Planning}, 2023.
\newblock URL \url{https://openreview.net/forum?id=9lkkqGagDF}.

\bibitem[Zheng et~al.(2023)Zheng, Wang, Xu, and Huang]{zheng2023is}
R.~Zheng, X.~Wang, H.~Xu, and F.~Huang.
\newblock Is model ensemble necessary? model-based {RL} via a single model with lipschitz regularized value function.
\newblock In \emph{The Eleventh International Conference on Learning Representations}, 2023.
\newblock URL \url{https://openreview.net/forum?id=hNyJBk3CwR}.

\bibitem[Du et~al.(2024)Du, Yang, Florence, Xia, Wahid, brian ichter, Sermanet, Yu, Abbeel, Tenenbaum, Kaelbling, Zeng, and Tompson]{du2024video}
Y.~Du, S.~Yang, P.~Florence, F.~Xia, A.~Wahid, brian ichter, P.~Sermanet, T.~Yu, P.~Abbeel, J.~B. Tenenbaum, L.~P. Kaelbling, A.~Zeng, and J.~Tompson.
\newblock Video language planning.
\newblock In \emph{The Twelfth International Conference on Learning Representations}, 2024.
\newblock URL \url{https://openreview.net/forum?id=9pKtcJcMP3}.

\bibitem[Huang et~al.(2025)Huang, Levy, Jiang, Anandkumar, Zhu, Fan, Huang, and Shrivastava]{huang2025ardupactiveregionvideo}
S.~Huang, M.~Levy, Z.~Jiang, A.~Anandkumar, Y.~Zhu, L.~Fan, D.-A. Huang, and A.~Shrivastava.
\newblock Ardup: Active region video diffusion for universal policies, 2025.
\newblock URL \url{https://arxiv.org/abs/2406.13301}.

\bibitem[Zhou et~al.(2024)Zhou, Pan, LeCun, and Pinto]{zhou2024dino}
G.~Zhou, H.~Pan, Y.~LeCun, and L.~Pinto.
\newblock Dino-wm: World models on pre-trained visual features enable zero-shot planning.
\newblock \emph{arXiv preprint arXiv:2411.04983}, 2024.

\bibitem[Schwarzer et~al.(2021{\natexlab{a}})Schwarzer, Rajkumar, Noukhovitch, Anand, Charlin, Hjelm, Bachman, and Courville]{pretrainspr2021}
M.~Schwarzer, N.~Rajkumar, M.~Noukhovitch, A.~Anand, L.~Charlin, R.~D. Hjelm, P.~Bachman, and A.~C. Courville.
\newblock Pretraining representations for data-efficient reinforcement learning.
\newblock In M.~Ranzato, A.~Beygelzimer, Y.~Dauphin, P.~Liang, and J.~W. Vaughan, editors, \emph{Advances in Neural Information Processing Systems}, volume~34, pages 12686--12699. Curran Associates, Inc., 2021{\natexlab{a}}.
\newblock URL \url{https://proceedings.neurips.cc/paper_files/paper/2021/file/69eba34671b3ef1ef38ee85caae6b2a1-Paper.pdf}.

\bibitem[Schwarzer et~al.(2021{\natexlab{b}})Schwarzer, Anand, Goel, Hjelm, Courville, and Bachman]{schwarzer2021spr}
M.~Schwarzer, A.~Anand, R.~Goel, R.~D. Hjelm, A.~Courville, and P.~Bachman.
\newblock Data-efficient reinforcement learning with self-predictive representations.
\newblock In \emph{International Conference on Learning Representations}, 2021{\natexlab{b}}.
\newblock URL \url{https://openreview.net/forum?id=uCQfPZwRaUu}.

\bibitem[Zheng et~al.(2023)Zheng, Wang, Sun, Ma, Zhao, Xu, Daum\'{e}~III, and Huang]{zheng2023taco}
R.~Zheng, X.~Wang, Y.~Sun, S.~Ma, J.~Zhao, H.~Xu, H.~Daum\'{e}~III, and F.~Huang.
\newblock Taco: Temporal latent action-driven contrastive loss for visual reinforcement learning.
\newblock In A.~Oh, T.~Naumann, A.~Globerson, K.~Saenko, M.~Hardt, and S.~Levine, editors, \emph{Advances in Neural Information Processing Systems}, volume~36, pages 48203--48225. Curran Associates, Inc., 2023.
\newblock URL \url{https://proceedings.neurips.cc/paper_files/paper/2023/file/96d00450ed65531ffe2996daed487536-Paper-Conference.pdf}.

\bibitem[Zheng et~al.(2024)Zheng, Liang, Wang, Ma, Daum\'{e}~III, Xu, Langford, Palanisamy, Basu, and Huang]{premiertaco2024}
R.~Zheng, Y.~Liang, X.~Wang, S.~Ma, H.~Daum\'{e}~III, H.~Xu, J.~Langford, P.~Palanisamy, K.~S. Basu, and F.~Huang.
\newblock Premier-taco is a few-shot policy learner: pretraining multitask representation via temporal action-driven contrastive loss.
\newblock In \emph{Proceedings of the 41st International Conference on Machine Learning}, ICML'24. JMLR.org, 2024.

\bibitem[Gao et~al.(2025)Gao, Gong, Guo, Hou, Lai, Li, Li, Lian, Liao, Liu, et~al.]{gao2025seedream}
Y.~Gao, L.~Gong, Q.~Guo, X.~Hou, Z.~Lai, F.~Li, L.~Li, X.~Lian, C.~Liao, L.~Liu, et~al.
\newblock Seedream 3.0 technical report.
\newblock \emph{arXiv preprint arXiv:2504.11346}, 2025.

\bibitem[Brohan et~al.(2022)Brohan, Brown, Carbajal, Chebotar, Dabis, Finn, Gopalakrishnan, Hausman, Herzog, Hsu, Ibarz, Ichter, Irpan, Jackson, Jesmonth, Joshi, Julian, Kalashnikov, Kuang, Leal, Lee, Levine, Lu, Malla, Manjunath, Mordatch, Nachum, Parada, Peralta, Perez, Pertsch, Quiambao, Rao, Ryoo, Salazar, Sanketi, Sayed, Singh, Sontakke, Stone, Tan, Tran, Vanhoucke, Vega, Vuong, Xia, Xiao, Xu, Xu, Yu, and Zitkovich]{rt1-2022}
A.~Brohan, N.~Brown, J.~Carbajal, Y.~Chebotar, J.~Dabis, C.~Finn, K.~Gopalakrishnan, K.~Hausman, A.~Herzog, J.~Hsu, J.~Ibarz, B.~Ichter, A.~Irpan, T.~Jackson, S.~Jesmonth, N.~Joshi, R.~Julian, D.~Kalashnikov, Y.~Kuang, I.~Leal, K.-H. Lee, S.~Levine, Y.~Lu, U.~Malla, D.~Manjunath, I.~Mordatch, O.~Nachum, C.~Parada, J.~Peralta, E.~Perez, K.~Pertsch, J.~Quiambao, K.~Rao, M.~Ryoo, G.~Salazar, P.~Sanketi, K.~Sayed, J.~Singh, S.~Sontakke, A.~Stone, C.~Tan, H.~Tran, V.~Vanhoucke, S.~Vega, Q.~Vuong, F.~Xia, T.~Xiao, P.~Xu, S.~Xu, T.~Yu, and B.~Zitkovich.
\newblock Rt-1: Robotics transformer for real-world control at scale.
\newblock In \emph{arXiv preprint arXiv:2212.06817}, 2022.

\bibitem[Brohan et~al.(2023)Brohan, Brown, Carbajal, Chebotar, Chen, Choromanski, Ding, Driess, Dubey, Finn, Florence, Fu, Arenas, Gopalakrishnan, Han, Hausman, Herzog, Hsu, Ichter, Irpan, Joshi, Julian, Kalashnikov, Kuang, Leal, Lee, Lee, Levine, Lu, Michalewski, Mordatch, Pertsch, Rao, Reymann, Ryoo, Salazar, Sanketi, Sermanet, Singh, Singh, Soricut, Tran, Vanhoucke, Vuong, Wahid, Welker, Wohlhart, Wu, Xia, Xiao, Xu, Xu, Yu, and Zitkovich]{rt22023arxiv}
A.~Brohan, N.~Brown, J.~Carbajal, Y.~Chebotar, X.~Chen, K.~Choromanski, T.~Ding, D.~Driess, A.~Dubey, C.~Finn, P.~Florence, C.~Fu, M.~G. Arenas, K.~Gopalakrishnan, K.~Han, K.~Hausman, A.~Herzog, J.~Hsu, B.~Ichter, A.~Irpan, N.~Joshi, R.~Julian, D.~Kalashnikov, Y.~Kuang, I.~Leal, L.~Lee, T.-W.~E. Lee, S.~Levine, Y.~Lu, H.~Michalewski, I.~Mordatch, K.~Pertsch, K.~Rao, K.~Reymann, M.~Ryoo, G.~Salazar, P.~Sanketi, P.~Sermanet, J.~Singh, A.~Singh, R.~Soricut, H.~Tran, V.~Vanhoucke, Q.~Vuong, A.~Wahid, S.~Welker, P.~Wohlhart, J.~Wu, F.~Xia, T.~Xiao, P.~Xu, S.~Xu, T.~Yu, and B.~Zitkovich.
\newblock Rt-2: Vision-language-action models transfer web knowledge to robotic control.
\newblock In \emph{arXiv preprint arXiv:2307.15818}, 2023.

\bibitem[Kim et~al.(2024)Kim, Pertsch, Karamcheti, Xiao, Balakrishna, Nair, Rafailov, Foster, Lam, Sanketi, Vuong, Kollar, Burchfiel, Tedrake, Sadigh, Levine, Liang, and Finn]{kim24openvla}
M.~Kim, K.~Pertsch, S.~Karamcheti, T.~Xiao, A.~Balakrishna, S.~Nair, R.~Rafailov, E.~Foster, G.~Lam, P.~Sanketi, Q.~Vuong, T.~Kollar, B.~Burchfiel, R.~Tedrake, D.~Sadigh, S.~Levine, P.~Liang, and C.~Finn.
\newblock Openvla: An open-source vision-language-action model.
\newblock \emph{arXiv preprint arXiv:2406.09246}, 2024.

\bibitem[Zheng et~al.(2025)Zheng, Liang, Huang, Gao, III, Kolobov, Huang, and Yang]{zheng2025tracevla}
R.~Zheng, Y.~Liang, S.~Huang, J.~Gao, H.~D. III, A.~Kolobov, F.~Huang, and J.~Yang.
\newblock Trace{VLA}: Visual trace prompting enhances spatial-temporal awareness for generalist robotic policies.
\newblock In \emph{The Thirteenth International Conference on Learning Representations}, 2025.

\bibitem[Wen et~al.(2024)Wen, Zhu, Li, Zhu, Wu, Xu, Cheng, Shen, Peng, Feng, et~al.]{wen2024tinyvla}
J.~Wen, Y.~Zhu, J.~Li, M.~Zhu, K.~Wu, Z.~Xu, R.~Cheng, C.~Shen, Y.~Peng, F.~Feng, et~al.
\newblock Tinyvla: Towards fast, data-efficient vision-language-action models for robotic manipulation.
\newblock \emph{arXiv preprint arXiv:2409.12514}, 2024.

\bibitem[Cheang et~al.(2024)Cheang, Chen, Jing, Kong, Li, Li, Liu, Wu, Xu, Yang, Zhang, and Zhu]{cheang2024gr2vla}
C.-L. Cheang, G.~Chen, Y.~Jing, T.~Kong, H.~Li, Y.~Li, Y.~Liu, H.~Wu, J.~Xu, Y.~Yang, H.~Zhang, and M.~Zhu.
\newblock Gr-2: A generative video-language-action model with web-scale knowledge for robot manipulation.
\newblock \emph{arXiv preprint arXiv:2410.06158}, 2024.

\bibitem[Li et~al.(2023)Li, Liu, Zhang, Yu, Xu, Wu, Cheang, Jing, Zhang, Liu, et~al.]{li2023vision}
X.~Li, M.~Liu, H.~Zhang, C.~Yu, J.~Xu, H.~Wu, C.~Cheang, Y.~Jing, W.~Zhang, H.~Liu, et~al.
\newblock Vision-language foundation models as effective robot imitators.
\newblock \emph{arXiv preprint arXiv:2311.01378}, 2023.

\bibitem[Zhen et~al.(2024)Zhen, Qiu, Chen, Yang, Yan, Du, Hong, and Gan]{zhen20243dvla}
H.~Zhen, X.~Qiu, P.~Chen, J.~Yang, X.~Yan, Y.~Du, Y.~Hong, and C.~Gan.
\newblock 3d-vla: 3d vision-language-action generative world model.
\newblock \emph{arXiv preprint arXiv:2403.09631}, 2024.

\bibitem[Huang et~al.(2024)Huang, Yong, Ma, Linghu, Li, Wang, Li, Zhu, Jia, and Huang]{huang2023embodied}
J.~Huang, S.~Yong, X.~Ma, X.~Linghu, P.~Li, Y.~Wang, Q.~Li, S.-C. Zhu, B.~Jia, and S.~Huang.
\newblock An embodied generalist agent in 3d world.
\newblock In \emph{Proceedings of the International Conference on Machine Learning (ICML)}, 2024.

\bibitem[Ye et~al.(2025)Ye, Jang, Jeon, Joo, Yang, Peng, Mandlekar, Tan, Chao, Lin, Liden, Lee, Gao, Zettlemoyer, Fox, and Seo]{ye2025latent}
S.~Ye, J.~Jang, B.~Jeon, S.~J. Joo, J.~Yang, B.~Peng, A.~Mandlekar, R.~Tan, Y.-W. Chao, B.~Y. Lin, L.~Liden, K.~Lee, J.~Gao, L.~Zettlemoyer, D.~Fox, and M.~Seo.
\newblock Latent action pretraining from videos.
\newblock In \emph{The Thirteenth International Conference on Learning Representations}, 2025.
\newblock URL \url{https://openreview.net/forum?id=VYOe2eBQeh}.

\bibitem[Yang et~al.(2025)Yang, Tan, Wu, Zheng, Peng, Liang, Gu, Cai, Ye, Jang, Deng, Liden, and Gao]{yang2025magma}
J.~Yang, R.~Tan, Q.~Wu, R.~Zheng, B.~Peng, Y.~Liang, Y.~Gu, M.~Cai, S.~Ye, J.~Jang, Y.~Deng, L.~Liden, and J.~Gao.
\newblock Magma: A foundation model for multimodal ai agents, 2025.
\newblock URL \url{https://arxiv.org/abs/2502.13130}.

\bibitem[Pertsch et~al.(2025)Pertsch, Stachowicz, Ichter, Driess, Nair, Vuong, Mees, Finn, and Levine]{pertsch2025fastefficientactiontokenization}
K.~Pertsch, K.~Stachowicz, B.~Ichter, D.~Driess, S.~Nair, Q.~Vuong, O.~Mees, C.~Finn, and S.~Levine.
\newblock Fast: Efficient action tokenization for vision-language-action models, 2025.
\newblock URL \url{https://arxiv.org/abs/2501.09747}.

\bibitem[Kim et~al.(2025)Kim, Finn, and Liang]{openvlaoft}
M.~J. Kim, C.~Finn, and P.~Liang.
\newblock Fine-tuning vision-language-action models: Optimizing speed and success.
\newblock \emph{arXiv preprint arXiv:2502.19645}, 2025.

\bibitem[{Octo Model Team} et~al.(2024){Octo Model Team}, Ghosh, Walke, Pertsch, Black, Mees, Dasari, Hejna, Xu, Luo, Kreiman, Tan, Chen, Sanketi, Vuong, Xiao, Sadigh, Finn, and Levine]{octo_2023}
{Octo Model Team}, D.~Ghosh, H.~Walke, K.~Pertsch, K.~Black, O.~Mees, S.~Dasari, J.~Hejna, C.~Xu, J.~Luo, T.~Kreiman, Y.~Tan, L.~Y. Chen, P.~Sanketi, Q.~Vuong, T.~Xiao, D.~Sadigh, C.~Finn, and S.~Levine.
\newblock Octo: An open-source generalist robot policy.
\newblock In \emph{Proceedings of Robotics: Science and Systems}, Delft, Netherlands, 2024.

\bibitem[Zeng et~al.(2024)Zeng, Bu, Wang, Xia, Chen, Dong, Song, Wang, Hu, Luo, et~al.]{zeng2024learning}
J.~Zeng, Q.~Bu, B.~Wang, W.~Xia, L.~Chen, H.~Dong, H.~Song, D.~Wang, D.~Hu, P.~Luo, et~al.
\newblock Learning manipulation by predicting interaction.
\newblock \emph{arXiv preprint arXiv:2406.00439}, 2024.

\bibitem[Bahl et~al.(2023)Bahl, Mendonca, Chen, Jain, and Pathak]{bahl2023affordances}
S.~Bahl, R.~Mendonca, L.~Chen, U.~Jain, and D.~Pathak.
\newblock Affordances from human videos as a versatile representation for robotics.
\newblock In \emph{Proceedings of the IEEE/CVF Conference on Computer Vision and Pattern Recognition}, 2023.

\bibitem[Kannan et~al.(2023)Kannan, Shaw, Bahl, Mannam, and Pathak]{kannan2023deft}
A.~Kannan, K.~Shaw, S.~Bahl, P.~Mannam, and D.~Pathak.
\newblock Deft: Dexterous fine-tuning for real-world hand policies.
\newblock \emph{arXiv preprint arXiv:2310.19797}, 2023.

\bibitem[Srirama et~al.(2024)Srirama, Dasari, Bahl, and Gupta]{srirama2024hrp}
M.~K. Srirama, S.~Dasari, S.~Bahl, and A.~Gupta.
\newblock Hrp: Human affordances for robotic pre-training.
\newblock \emph{arXiv preprint arXiv:2407.18911}, 2024.

\bibitem[Shaw et~al.(2023)Shaw, Bahl, and Pathak]{shaw2023videodex}
K.~Shaw, S.~Bahl, and D.~Pathak.
\newblock Videodex: Learning dexterity from internet videos.
\newblock In \emph{Conference on Robot Learning}, 2023.

\bibitem[Wen et~al.(2023)Wen, Lin, So, Chen, Dou, Gao, and Abbeel]{wen2023any}
C.~Wen, X.~Lin, J.~So, K.~Chen, Q.~Dou, Y.~Gao, and P.~Abbeel.
\newblock Any-point trajectory modeling for policy learning.
\newblock \emph{arXiv preprint arXiv:2401.00025}, 2023.

\bibitem[Bharadhwaj et~al.(2024)Bharadhwaj, Mottaghi, Gupta, and Tulsiani]{bharadhwaj2024track2act}
H.~Bharadhwaj, R.~Mottaghi, A.~Gupta, and S.~Tulsiani.
\newblock Track2act: Predicting point tracks from internet videos enables diverse zero-shot robot manipulation.
\newblock \emph{arXiv e-prints}, pages arXiv--2405, 2024.

\bibitem[Wang et~al.(2023)Wang, Fan, Sun, Zhang, Fei-Fei, Xu, Zhu, and Anandkumar]{wang2023mimicplay}
C.~Wang, L.~Fan, J.~Sun, R.~Zhang, L.~Fei-Fei, D.~Xu, Y.~Zhu, and A.~Anandkumar.
\newblock Mimicplay: Long-horizon imitation learning by watching human play.
\newblock \emph{arXiv preprint arXiv:2302.12422}, 2023.

\bibitem[Zhu et~al.(2024)Zhu, Lim, Stone, and Zhu]{zhu2024vision}
Y.~Zhu, A.~Lim, P.~Stone, and Y.~Zhu.
\newblock Vision-based manipulation from single human video with open-world object graphs.
\newblock \emph{arXiv preprint arXiv:2405.20321}, 2024.

\bibitem[Bharadhwaj et~al.(2023)Bharadhwaj, Gupta, Tulsiani, and Kumar]{bharadhwaj2023zero}
H.~Bharadhwaj, A.~Gupta, S.~Tulsiani, and V.~Kumar.
\newblock Zero-shot robot manipulation from passive human videos.
\newblock \emph{arXiv preprint arXiv:2302.02011}, 2023.

\bibitem[Ye et~al.(2023)Ye, Wang, Huang, Qin, and Wang]{ye2023learning}
J.~Ye, J.~Wang, B.~Huang, Y.~Qin, and X.~Wang.
\newblock Learning continuous grasping function with a dexterous hand from human demonstrations.
\newblock \emph{IEEE Robotics and Automation Letters}, 8\penalty0 (5):\penalty0 2882--2889, 2023.

\bibitem[Qin et~al.(2022)Qin, Wu, Liu, Jiang, Yang, Fu, and Wang]{qin2022dexmv}
Y.~Qin, Y.-H. Wu, S.~Liu, H.~Jiang, R.~Yang, Y.~Fu, and X.~Wang.
\newblock Dexmv: Imitation learning for dexterous manipulation from human videos.
\newblock In \emph{European Conference on Computer Vision}, 2022.

\bibitem[Yang et~al.(2024)Yang, Cao, Deng, Antonova, Song, and Bohg]{yang2024equibot}
J.~Yang, Z.-a. Cao, C.~Deng, R.~Antonova, S.~Song, and J.~Bohg.
\newblock Equibot: Sim (3)-equivariant diffusion policy for generalizable and data efficient learning.
\newblock \emph{arXiv preprint arXiv:2407.01479}, 2024.

\bibitem[Bruce et~al.(2024)Bruce, Dennis, Edwards, Parker-Holder, Shi, Hughes, Lai, Mavalankar, Steigerwald, Apps, Aytar, Bechtle, Behbahani, Chan, Heess, Gonzalez, Osindero, Ozair, Reed, Zhang, Zolna, Clune, de~Freitas, Singh, and Rocktäschel]{bruce2024geniegenerativeinteractiveenvironments}
J.~Bruce, M.~Dennis, A.~Edwards, J.~Parker-Holder, Y.~Shi, E.~Hughes, M.~Lai, A.~Mavalankar, R.~Steigerwald, C.~Apps, Y.~Aytar, S.~Bechtle, F.~Behbahani, S.~Chan, N.~Heess, L.~Gonzalez, S.~Osindero, S.~Ozair, S.~Reed, J.~Zhang, K.~Zolna, J.~Clune, N.~de~Freitas, S.~Singh, and T.~Rocktäschel.
\newblock Genie: Generative interactive environments, 2024.
\newblock URL \url{https://arxiv.org/abs/2402.15391}.

\bibitem[Chen et~al.(2025)Chen, Ge, Tang, Li, Ge, Ding, Shan, and Liu]{chen2025motolatentmotiontoken}
Y.~Chen, Y.~Ge, W.~Tang, Y.~Li, Y.~Ge, M.~Ding, Y.~Shan, and X.~Liu.
\newblock Moto: Latent motion token as the bridging language for learning robot manipulation from videos, 2025.
\newblock URL \url{https://arxiv.org/abs/2412.04445}.

\bibitem[Schmidt and Jiang(2024)]{schmidt2024learning}
D.~Schmidt and M.~Jiang.
\newblock Learning to act without actions.
\newblock In \emph{The Twelfth International Conference on Learning Representations}, 2024.
\newblock URL \url{https://openreview.net/forum?id=rvUq3cxpDF}.

\bibitem[Ren et~al.(2025)Ren, Wei, Guo, Zhao, Kang, Feng, and Jin]{ren2025videoworldexploringknowledgelearning}
Z.~Ren, Y.~Wei, X.~Guo, Y.~Zhao, B.~Kang, J.~Feng, and X.~Jin.
\newblock Videoworld: Exploring knowledge learning from unlabeled videos, 2025.
\newblock URL \url{https://arxiv.org/abs/2501.09781}.

\bibitem[Bu et~al.(2025)Bu, Cai, Chen, Cui, Ding, Feng, Gao, He, Huang, Jiang, et~al.]{bu2025agibot}
Q.~Bu, J.~Cai, L.~Chen, X.~Cui, Y.~Ding, S.~Feng, S.~Gao, X.~He, X.~Huang, S.~Jiang, et~al.
\newblock Agibot world colosseo: A large-scale manipulation platform for scalable and intelligent embodied systems.
\newblock \emph{arXiv preprint arXiv:2503.06669}, 2025.

\bibitem[Khazatsky et~al.(2024)Khazatsky, Pertsch, Nair, Balakrishna, Dasari, Karamcheti, Nasiriany, Srirama, Chen, Ellis, Fagan, Hejna, Itkina, Lepert, Ma, Miller, Wu, Belkhale, Dass, Ha, Jain, Lee, Lee, Memmel, Park, Radosavovic, Wang, Zhan, Black, Chi, Hatch, Lin, Lu, Mercat, Rehman, Sanketi, Sharma, Simpson, Vuong, Walke, Wulfe, Xiao, Yang, Yavary, Zhao, Agia, Baijal, Castro, Chen, Chen, Chung, Drake, Foster, Gao, Herrera, Heo, Hsu, Hu, Jackson, Le, Li, Lin, Lin, Ma, Maddukuri, Mirchandani, Morton, Nguyen, O'Neill, Scalise, Seale, Son, Tian, Tran, Wang, Wu, Xie, Yang, Yin, Zhang, Bastani, Berseth, Bohg, Goldberg, Gupta, Gupta, Jayaraman, Lim, Malik, Martín-Martín, Ramamoorthy, Sadigh, Song, Wu, Yip, Zhu, Kollar, Levine, and Finn]{khazatsky2024droid}
A.~Khazatsky, K.~Pertsch, S.~Nair, A.~Balakrishna, S.~Dasari, S.~Karamcheti, S.~Nasiriany, M.~K. Srirama, L.~Y. Chen, K.~Ellis, P.~D. Fagan, J.~Hejna, M.~Itkina, M.~Lepert, Y.~J. Ma, P.~T. Miller, J.~Wu, S.~Belkhale, S.~Dass, H.~Ha, A.~Jain, A.~Lee, Y.~Lee, M.~Memmel, S.~Park, I.~Radosavovic, K.~Wang, A.~Zhan, K.~Black, C.~Chi, K.~B. Hatch, S.~Lin, J.~Lu, J.~Mercat, A.~Rehman, P.~R. Sanketi, A.~Sharma, C.~Simpson, Q.~Vuong, H.~R. Walke, B.~Wulfe, T.~Xiao, J.~H. Yang, A.~Yavary, T.~Z. Zhao, C.~Agia, R.~Baijal, M.~G. Castro, D.~Chen, Q.~Chen, T.~Chung, J.~Drake, E.~P. Foster, J.~Gao, D.~A. Herrera, M.~Heo, K.~Hsu, J.~Hu, D.~Jackson, C.~Le, Y.~Li, K.~Lin, R.~Lin, Z.~Ma, A.~Maddukuri, S.~Mirchandani, D.~Morton, T.~Nguyen, A.~O'Neill, R.~Scalise, D.~Seale, V.~Son, S.~Tian, E.~Tran, A.~E. Wang, Y.~Wu, A.~Xie, J.~Yang, P.~Yin, Y.~Zhang, O.~Bastani, G.~Berseth, J.~Bohg, K.~Goldberg, A.~Gupta, A.~Gupta, D.~Jayaraman, J.~J. Lim, J.~Malik, R.~Martín-Martín, S.~Ramamoorthy, D.~Sadigh, S.~Song, J.~Wu, M.~C. Yip, Y.~Zhu,
  T.~Kollar, S.~Levine, and C.~Finn.
\newblock Droid: A large-scale in-the-wild robot manipulation dataset.
\newblock 2024.

\bibitem[Lynch et~al.(2022)Lynch, Wahid, Tompson, Ding, Betker, Baruch, Armstrong, and Florence]{lynch2022interactivelanguagetalkingrobots}
C.~Lynch, A.~Wahid, J.~Tompson, T.~Ding, J.~Betker, R.~Baruch, T.~Armstrong, and P.~Florence.
\newblock Interactive language: Talking to robots in real time, 2022.

\bibitem[Walke et~al.(2023)Walke, Black, Lee, Kim, Du, Zheng, Zhao, Hansen-Estruch, Vuong, He, Myers, Fang, Finn, and Levine]{walke2023bridgedata}
H.~Walke, K.~Black, A.~Lee, M.~J. Kim, M.~Du, C.~Zheng, T.~Zhao, P.~Hansen-Estruch, Q.~Vuong, A.~He, V.~Myers, K.~Fang, C.~Finn, and S.~Levine.
\newblock Bridgedata v2: A dataset for robot learning at scale.
\newblock In \emph{Conference on Robot Learning (CoRL)}, 2023.

\bibitem[Shah et~al.(2023)Shah, Mart{\'\i}n-Mart{\'\i}n, and Zhu]{shah2023mutex}
R.~Shah, R.~Mart{\'\i}n-Mart{\'\i}n, and Y.~Zhu.
\newblock Mutex: Learning unified policies from multimodal task specifications.
\newblock In \emph{7th Annual Conference on Robot Learning}, 2023.

\bibitem[Thomas et~al.(2023)Thomas, Cheng, Loynd, Frujeri, Vineet, Jalobeanu, and Kolobov]{thomas2023plex}
G.~Thomas, C.-A. Cheng, R.~Loynd, F.~V. Frujeri, V.~Vineet, M.~Jalobeanu, and A.~Kolobov.
\newblock Plex: Making the most of the available data for robotic manipulation pretraining.
\newblock In \emph{CoRL}, 2023.

\bibitem[Bharadhwaj et~al.(2024)Bharadhwaj, Vakil, Sharma, Gupta, Tulsiani, and Kumar]{roboset}
H.~Bharadhwaj, J.~Vakil, M.~Sharma, A.~Gupta, S.~Tulsiani, and V.~Kumar.
\newblock Roboagent: Generalization and efficiency in robot manipulation via semantic augmentations and action chunking.
\newblock In \emph{2024 IEEE International Conference on Robotics and Automation (ICRA)}, 2024.

\bibitem[Loshchilov and Hutter(2019)]{adamw}
I.~Loshchilov and F.~Hutter.
\newblock Decoupled weight decay regularization.
\newblock In \emph{International Conference on Learning Representations}, 2019.
\newblock URL \url{https://openreview.net/forum?id=Bkg6RiCqY7}.

\end{thebibliography}

\end{document}